\pdfoutput=1
\documentclass[11pt]{article}

\usepackage{graphicx}
\usepackage{multirow}
\usepackage{booktabs}
\usepackage{amssymb}
\usepackage{amsmath}
\usepackage{xcolor}

\usepackage[final]{acl}

\usepackage{times}
\usepackage{latexsym}

\usepackage[T1]{fontenc}
\usepackage[utf8]{inputenc}

\usepackage{microtype}

\usepackage{inconsolata}

\usepackage{graphicx}

\title{DiaHalu: A Dialogue-level Hallucination Evaluation Benchmark \\ for Large Language Models}

\author{Kedi Chen, Qin Chen\thanks{Corresponding author.}, Jie Zhou, Yishen He \and Liang He \\
        School of Computer Science and Technology, East China Normal University, Shanghai, China \\
        \{kdchen,10215102502\}@stu.ecnu.edu.cn \{qchen, jzhou, lhe\}@cs.ecnu.edu.cn\\}
   
\begin{document}
\maketitle
\begin{abstract}
Though large language models (LLMs) achieve significant success in recent years, the hallucination issue remains a challenge, and numerous benchmarks are proposed for hallucination detection. Nevertheless, some of these benchmarks are not naturally generated by LLMs but are intentionally induced. Also, many merely focus on the factuality hallucination while ignoring the faithfulness hallucination. Additionally, although dialogue pattern is more widely utilized in the era of LLMs, current benchmarks only concentrate on sentence-level and passage-level hallucination. In this study, we propose DiaHalu, the first dedicated dialogue-level hallucination evaluation benchmark for LLMs to our knowledge. Initially, we integrate the collected topics into system prompts and facilitate a dialogue between two LLMs. Subsequently, we manually modify the contents that do not adhere to human language conventions and then have LLMs re-generate, simulating authentic human-machine interaction scenarios. Finally, professional scholars annotate all the samples in the dataset. DiaHalu covers four common multi-turn dialogue domains and five hallucination subtypes, extended from factuality and faithfulness hallucination. Experiments with the well-known LLMs and detection methods show that DiaHalu is a challenging benchmark, holding significant values for further research\footnote{https://github.com/ECNU-ICALK/DiaHalu}.
\end{abstract}

\section{Introduction}

\begin{figure}[!ht]
    \centering
    \includegraphics[scale=0.30]{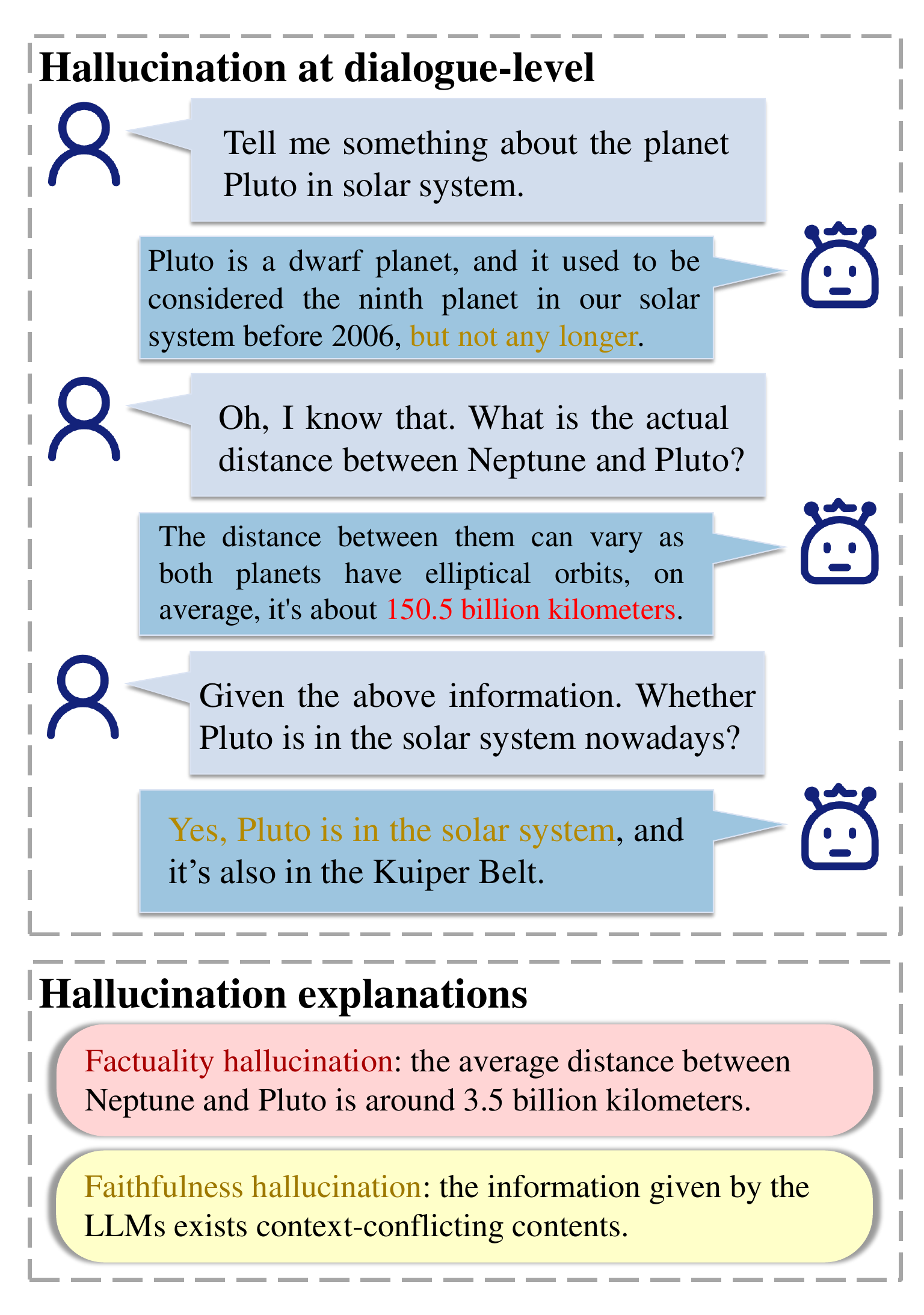}
    \caption{Our benchmark not only includes factuality hallucination but also incorporates faithfulness hallucination at the dialogue level, although most benchmarks overlook the latter one.}
    \label{fig: fac and faith hallu}
\end{figure}

Large language models (LLMs) \citep{DBLP:journals/corr/abs-2303-18223}, addressing many downstream tasks through natural language generation (NLG) technique, achieve significant success across diverse fields of natural language processing (NLP) \citep{li2024leveraging,10387715,DBLP:journals/corr/abs-2302-10205,DBLP:journals/corr/abs-2307-02046}. With a considerable volume of parameters and sophisticated training methodologies, LLMs significantly propelled advancements in artificial intelligence \citep{DBLP:journals/corr/abs-2303-18223}.

Despite many advantages of LLMs, the issue of hallucination remains a primary concern \citep{DBLP:journals/csur/JiLFYSXIBMF23,DBLP:journals/corr/abs-2309-01219}. Hallucination predominantly centers on the inclination of LLMs to generate nonsensical or untruthful contents for specific sources \citep{DBLP:journals/corr/abs-2310-07521}. Therefore, the occurrence of hallucination poses risks to the application of LLMs in various real-world scenarios \citep{DBLP:journals/corr/abs-2309-00087,DBLP:journals/corr/abs-2303-17564,DBLP:conf/coling/Chen0C0024}.
% such as medicine \citep{DBLP:journals/corr/abs-2309-00087}, education \citep{kasneci2023chatgpt} and finance \citep{DBLP:journals/corr/abs-2303-17564} etc. 

Given the aforementioned risks, hallucination detection emerges as a highly crucial task. In recent years, researchers propose numerous benchmarks for hallucination detection task \citep{DBLP:conf/emnlp/LiCZNW23, DBLP:journals/corr/abs-2310-14564, DBLP:conf/emnlp/ManakulLG23, DBLP:conf/emnlp/YangS023}. Nevertheless, several problems persist in these benchmarks. \textbf{(1) Not naturally generated.} One pitfall of existing benchmarks for detecting LLMs' hallucination is that the hallucinated contents are typically induced via manually designed trigger prompts \citep{DBLP:conf/emnlp/LiCZNW23}, while not naturally generated by LLMs as in daily usage \citep{DBLP:conf/acl/LiuZBMSCD22}. \textbf{(2) Merely focusing on factuality hallucination. } Most benchmarks merely focus on detecting factuality hallucination \citep{DBLP:journals/corr/abs-2310-14564}, with few datasets that can demonstrate faithfulness hallucination \citep{DBLP:journals/corr/abs-2311-05232} (Figure~\ref{fig: fac and faith hallu}). Factuality hallucination emphasizes the discrepancy between generated contents and real-world facts. Faithfulness hallucination refers to the divergence of generated contents from user instructions or other inputs, as well as self-consistency within the generated contents, which tends to be ignored. \textbf{(3) Only concentrating on sentence-level and passage-level.} Researchers propose many sentence-level \citep{DBLP:conf/emnlp/ManakulLG23,DBLP:journals/corr/abs-2310-17918} and passage-level \citep{DBLP:conf/emnlp/YangS023,DBLP:conf/emnlp/FengBBT23} hallucination detection benchmarks. However, the dialogue pattern has broader and more widespread applications in LLMs. More unique types of hallucination in dialogues make them more challenging to be detected (detailed explanations in Appendix~\ref{app: future works}). Although it is rarely mentioned in previous researches, dialogue-level hallucination detection is equally essential.

Therefore, we propose a new \textbf{dia}logue-level \textbf{hal}l\textbf{u}cination evaluation benchmark for large language models (\textbf{DiaHalu}). We \textbf{initially} determine four domains for multi-turn dialogue: knowledge-grounded, task-oriented, chit-chat and reasoning. For each domain, we undertake a three-step process to construct the dataset. (1) We collect topics for dialogue from various sources, incorporate the topics into artificially designed system prompts and input them into two LLMs, enabling them to engage in a multi-turn dialogue. (2) Since the knowledge-grounded and task-oriented domains stand for real human-machine interaction scenarios, we align the contents of one of the conversational participants with human language. We manually modify the contents that do not conform to human language conventions and have LLMs re-generate, resulting in the final responses. (3) Professional scholars annotate all the samples with labels, hallucination subtypes and locations, as well as explanations. It is noteworthy that we not only consider the factuality hallucination but also further classify the faithfulness hallucination into three types: Incoherence, Irrelevance and Overreliance. We similarly introduce the reasoning hallucination for the reasoning domain. The advantages of ours compared with previous benchmarks are listed in Table~\ref{tab: comparison}. \textbf{Additionally}, we conduct experiments on DiaHalu by deploying existing hallucination detection methods and some commonly used LLMs. The results indicate that DiaHalu is a highly challenging benchmark.
% and both Chain-of-Thought and retrieval methods can improve the performance of the detection. 
Our contributions can be listed as follows:
\begin{itemize}
\item{ To our current knowledge, we propose the first \textit{dedicated} dialogue-level hallucination detection benchmark for LLMs named DiaHalu.}
%\footnote{https://anonymous.4open.science/r/DiaHalu-A1E9/}
\item{DiaHalu covers four multi-turn dialogue domains along with five hallucination subtypes extended from factuality and faithfulness hallucination, which are more widely applicable in real-world scenarios.}
\item{The experimental results indicate that DiaHalu is a highly challenging benchmark for most LLMs and existing detection methods, holding significant value for further researches.}
\end{itemize}

\begin{table}[!t]
\centering
\fontsize{7pt}{8.4pt}\selectfont
\renewcommand{\arraystretch}{1.3} 
\begin{tabular}{c|cccc}
\toprule
\textbf{Benchmark}  & \textbf{By LLMs} & \textbf{Faith Halu} &  \textbf{Multi Dia} & \textbf{Explanation} \\
\midrule
FactCollect  & - & - & - & -           \\
BEGIN & - & - & $\checkmark$ & $\checkmark$         \\
HADES & - & $\checkmark$ & - & $\checkmark$\\
FactCHD & $\checkmark$ & - & - & $\checkmark$\\
HaluEval & - & - & $\checkmark$ & $\checkmark$ \\
WikiBio+ & $\checkmark$ & - & - & - \\
PHD & $\checkmark$ & - & - & - \\
\midrule
Ours & $\checkmark$ & $\checkmark$ &  $\checkmark$ & $\checkmark$ \\
\bottomrule
\end{tabular}
\caption{The comparison between our DiaHalu and other benchmarks. `By LLMs', `Faith Halu', `Multi Dia', and `Explanation' mean whether it is naturally generated by LLMs, whether it provides faithfulness hallucination, whether it is at multi-turn dialogue level, and whether there are explanations respectively (Appendix~\ref{app: comparison}).}
\label{tab: comparison}
\end{table}

\section{Related Work}
\subsection{Hallucination Detection Benchmarks}
% Hallucination predominantly centers on the inclination of large language models to generate nonsensical or untruthful contents for specific sources \citep{DBLP:journals/corr/abs-2310-07521}. Therefore, detecting hallucination in LLMs is of great importance. 

In recent years, researchers propose numerous benchmarks for hallucination detection.

In earlier years, hallucination detection benchmarks are primarily organized through manual methods or generated via conventional language models. FactCollect \citep{DBLP:conf/naacl/RibeiroLGDB22} is an artificially generated, multi-source factual hallucination detection benchmark. \citet{DBLP:journals/corr/abs-2307-06908} collects error samples by instructing the language model based on pre-defined error types. HADES \citep{DBLP:conf/acl/LiuZBMSCD22} and BEGIN \citep{DBLP:journals/tacl/DziriRLR22} constitute hallucination detection datasets by conventional language model BERT \citep{DBLP:conf/naacl/DevlinCLT19} and T5 (mostly) \citep{2020t5} respectively. These benchmarks are not naturally generated by LLMs as in daily usage.

Consequently, some benchmarks are proposed to investigate the direct generation abilities of large language models. \citet{DBLP:journals/corr/abs-2310-17918,DBLP:journals/corr/abs-2311-00681,DBLP:conf/emnlp/ChenDBQWCW23,DBLP:journals/corr/abs-2307-10236,zheng2023does} enable LLMs to handle Question-Answer (QA) task and assess the factual accuracy of their responses. Concept-7 dataset used by \citet{DBLP:journals/corr/abs-2309-02654} evaluates whether a language model truly comprehends the meaning of each concept, thereby determining the presence of hallucination. FactCHD \citep{DBLP:journals/corr/abs-2310-12086} is generated based on natural language text and knowledge graphs (KGs).
% deploying two major steps: Query and Response Data Simulation \& Chain of Evidence and Human Filter, to detect fact-conflicting hallucination in LLMs.
\citet{mündler2023selfcontradictory} employs a generative language model (gLM) to rewrite sentences according to the given context. New sentences compose a dataset that can evaluate whether the generated sentences exhibit knowledge-based self-contradiction hallucination. The aforementioned benchmarks mainly focus on detecting factuality hallucination \citep{DBLP:journals/corr/abs-2310-14564}, while ignoring the faithfulness hallucination \citep{DBLP:journals/corr/abs-2311-05232}. 
The benchmark proposed in this paper extends to include faithfulness hallucination, that is, to evaluate the coherence and relevance of contents generated by LLMs.
 
Researchers also raise many sentence-level \citep{DBLP:conf/emnlp/ManakulLG23,DBLP:journals/corr/abs-2310-17918,DBLP:journals/corr/abs-2310-14564} and passage-level \citep{DBLP:conf/emnlp/YangS023,DBLP:conf/emnlp/FengBBT23,DBLP:conf/emnlp/LiCZNW23} hallucination detection benchmarks. Nevertheless, the dialogue pattern holds broader applications within LLMs. While previous researches rarely, to our current knowledge, propose a dialogue-level hallucination detection benchmark for LLMs. So, our DiaHalu is at the dialogue level.

% \subsection{Dialogue System in LLMs}
% It is the formidable dialogue capability exhibited by ChatGPT3.5 \citep{10113601} and GPT4 \citep{openai2023gpt4}, that stimulates a heightened interest among researchers in large language models. 
% The ability for multi-turn dialogue is among the most crucial features of LLMs, which has broader and more widespread applications in the current development of NLP. 

% Dialogue system can be primarily categorized into various types, including task-oriented dialogue \citep{DBLP:conf/eacl/Rojas-BarahonaG17}, chit-chat dialogue \citep{DBLP:conf/naacl/SunMCRSLWLCC21}, knowledge-grounded dialogue \citep{DBLP:conf/aaai/GhazvininejadBC18}, etc. In these scenarios, language models are required to exhibit long-term memory, recognize shifts in dialogue states, encompass extensive knowledge, and engage in reasoning processes \citep{DBLP:conf/acl/0009C23}. Despite the remarkable general capability of LLMs, there are still deficiencies in handling complex dialogue scenarios \citep{DBLP:journals/sigkdd/ChenLYT17,DBLP:journals/air/DeriuROERAC21}.

% Due to the critical significance of the dialogue ability, this paper underscores the detection of hallucination produced by LLMs in multi-turn dialogues.

\subsection{Hallucination Detection}
Current methods for hallucination detection \citep{DBLP:journals/corr/abs-2401-01313} can mainly be divided into four categories. (1) Model-based. This method involves having the language models perform a classification task to determine whether hallucinated contents are present \citep{DBLP:conf/emnlp/ZhaoND23,DBLP:conf/emnlp/MaharajSKMB23}. (2) Retrieval-based. For the limited knowledge within the parameters of language models, we can detect hallucination by extracting or retrieving relevant knowledge from external knowledge graphs \citep{DBLP:conf/esws/MartinoIT23,DBLP:journals/corr/abs-2310-12086} or web information sources \citep{DBLP:journals/corr/abs-2404-08189}. (3) Sampling-based. Another feasible method is to rewrite the generated contents to evaluate the consistency \citep{DBLP:conf/emnlp/ManakulLG23,DBLP:journals/corr/abs-2310-17918,DBLP:journals/corr/abs-2311-01740}. (4) Uncertainty-based. The mainstream view of this method \citep{DBLP:conf/emnlp/ZhangQGDZZZWF23,DBLP:journals/corr/abs-2404-10136} is that `the lower the probability of generating a token, the more likely a model is to produce hallucination'.

\begin{figure*}[!t]
    \centering
    \includegraphics[width=480pt,height=273pt]{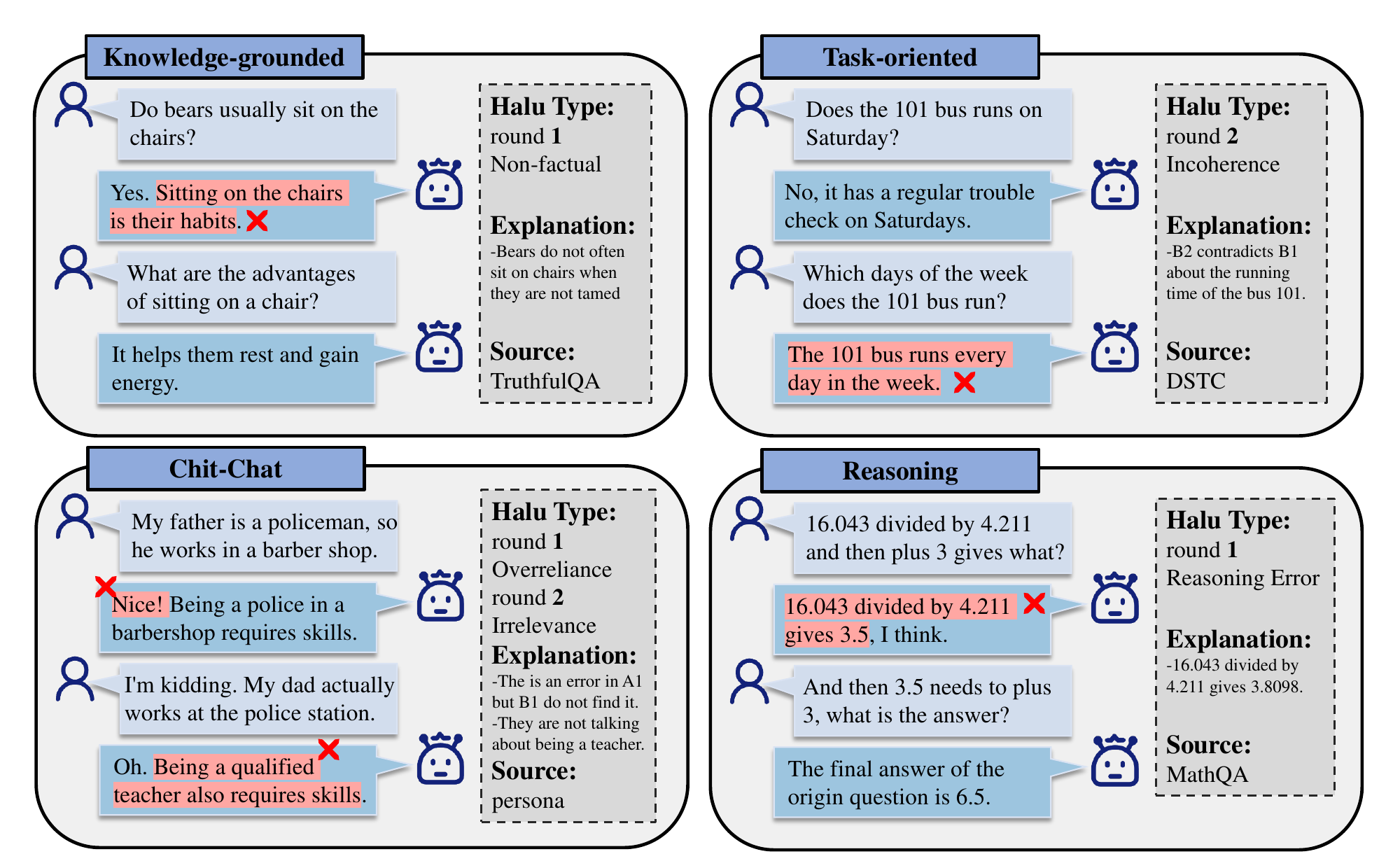}
    \caption{The demonstration of the DiaHalu benchmark, which covers four domains and five hallucination subtypes within dialogue-level scenarios. We also provide explanations and sources in the benchmark. }
    \label{fig: four domains and five types}
\end{figure*}

\section{The Overview of DiaHalu}
\subsection{Principles}
The primary objective of this benchmark is to conduct hallucination detection in large language models. Hence, it is imperative to comprehend the meaning of \textit{hallucination}. Hallucination predominantly centers on the inclination of LLMs to generate nonsensical or untruthful contents regarding specific sources \citep{DBLP:journals/corr/abs-2310-07521}. The significance of hallucination detection lies in elevating the quality of text generation, preventing misleading information and misunderstandings, supporting applications within professional domains, etc. Therefore, to enhance the universality of our benchmark, it encompasses various multi-turn dialogue scenarios and multiple subtypes of hallucination, extending from factuality hallucination and faithfulness hallucination \citep{DBLP:journals/corr/abs-2311-05232}.

\subsection{Hallucination on Diverse Domains}
We consider hallucination in diverse domains of multi-turn dialogue scenarios.
Our benchmark covers a total of four domains, shown in Figure~\ref{fig: four domains and five types}. Their specific descriptions are in Appendix~\ref{app:four domains}.

\paragraph{Knowledge-grounded dialogue} is designed for users to engage in knowledge-based dialogue with LLMs \citep{DBLP:conf/aaai/GhazvininejadBC18}. The two speakers take part in a conversation about a knowledge-based issue.

% It principally examines the accuracy of knowledge of the parameters in LLMs \citep{DBLP:conf/emnlp/PetroniRRLBWM19}.

\paragraph{Task-oriented dialogue} is in a form of human-computer interaction, intending to accomplish a user-specified task \citep{DBLP:conf/eacl/Rojas-BarahonaG17}. 

\paragraph{Chit-Chat dialogue} involves open-ended and non-goal dialogue \citep{DBLP:conf/naacl/SunMCRSLWLCC21}. 
We provide two LLMs with personas and facilitate a dialogue between them. 

\paragraph{Reasoning dialogue} Following previous works \citep{DBLP:journals/corr/abs-2310-00741,DBLP:journals/corr/abs-2311-09702,DBLP:journals/corr/abs-2401-00290,DBLP:journals/corr/abs-2402-19405,zheng2023does,DBLP:conf/acl/0009C23}, we also treat reasoning errors as a kind of hallucination.
We have the models discuss mathematical problems to achieve the answers \citep{DBLP:journals/corr/abs-2401-03238}.

\subsection{Hallucination Taxonomy}
We consider both factuality and faithfulness hallucination \citep{DBLP:journals/corr/abs-2311-05232}. 
Based on \citet{DBLP:journals/corr/abs-2310-00741,DBLP:conf/emnlp/WuSLMBW23,DBLP:conf/naacl/DziriMYZR22} and early works on text coherence \citep{DBLP:conf/coling/WolfG04,DBLP:conf/coling/AtwellISA24}, we carry out a detailed classification of faithfulness hallucination into Incoherence, Irrelevance, and Overreliance. Meanwhile, we introduce Reasoning Error within the reasoning dialogue \citep{DBLP:journals/corr/abs-2310-00741,DBLP:journals/corr/abs-2311-09702,DBLP:journals/corr/abs-2401-00290,DBLP:journals/corr/abs-2402-19405,zheng2023does,DBLP:conf/acl/0009C23}. The example of each type can be referenced in Figure~\ref{fig: four domains and five types}. 

% and Appendix~\ref{app: application scope of hallucination labels}.

\paragraph{Non-factual} implies whether it aligns with factual information.

\paragraph{Incoherence} includes input-conflicting, context-conflicting and self-conflicting contents in the dialogue.

\paragraph{Irrelevance} means that something unrelated to the topic of the conversation comes up.

\paragraph{Overreliance} is that the LLM excessively trusts in the correctness of the context, generating responses for unanswerable contents \citep{DBLP:journals/corr/abs-2310-11877}.

\paragraph{Reasoning Error} covers all errors within the reasoning dialogue.

\section{The Construction of DiaHalu}
\subsection{The Collection of Dialogue Topics}
Since we confirm four domains for DiaHalu, the first step is to collect the topics for each dialogic domain.

For \textbf{knowledge-grounded dialogue}, we take into account world knowledge, factual knowledge, commonsense knowledge and multi-hop web knowledge. Therefore, we gather dialogue topics from TruthfulQA \citep{DBLP:conf/acl/LinHE22}, CommonsenseQA \citep{DBLP:conf/naacl/TalmorHLB19} and CWQ \citep{DBLP:conf/naacl/TalmorB18} datasets. There are also topics provided by GPT4 \citep{openai2023gpt4} and social media (including the authors). As for \textbf{task-oriented dialogue}, we primarily apply the most widely used MultiWOZ (MultiWOZ 2.1) \citep{DBLP:conf/emnlp/BudzianowskiWTC18} which covers 7 real-life scenarios. To enrich the dialogue settings, we also consider the DSTC (DSTC 1.0) \citep{DBLP:conf/sigdial/WilliamsRRB13} dataset with a focus on bus routes. GPT4 and social media are harnessed to augment user behaviors and generate more dialogue occasions. We define the LLMs with personas primarily from \citet{DBLP:journals/corr/abs-2312-10007} and facilitate an open \textbf{chit-chat dialogue} between them. Additionally, we make use of mathematical problems to assess the logical \textbf{reasoning dialogue} abilities of LLMs. These problems are sourced from GSM8K \citep{DBLP:journals/corr/abs-2110-14168} and MathQA \citep{amini2019mathqa}, both of which involve mathematical problems and solving processes encountered by middle school students. 

The overall distribution of the above sources for dialogue topics is illustrated in Appendix~\ref{app: distribution}.

\subsection{Dialogue Generation}
\begin{figure}[!t]
    \centering
    \includegraphics[scale=0.27]{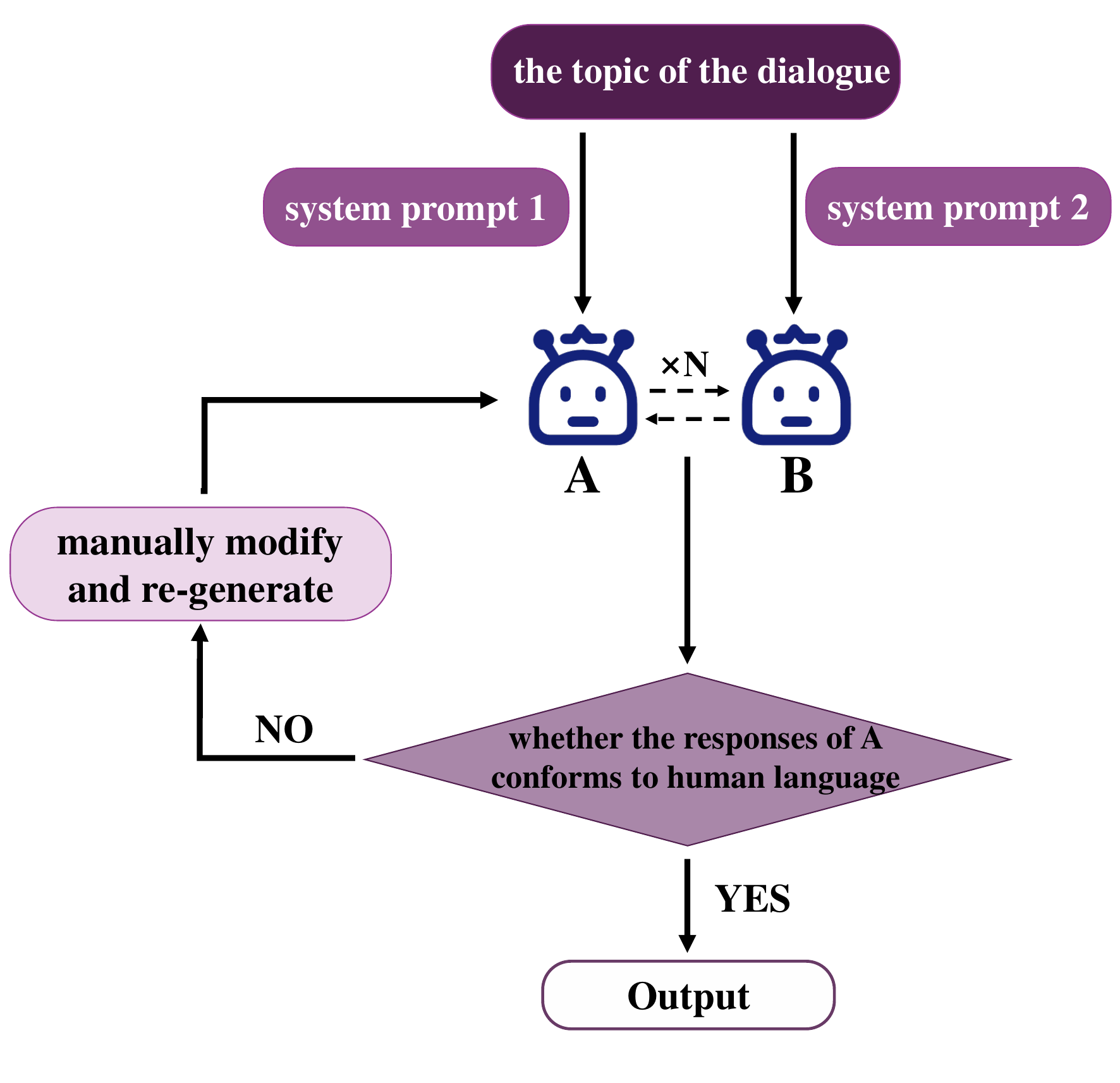}
    \caption{The complete process of dialogue generation.}
    \label{fig: dialogue generation}
\end{figure}

Once finishing collecting the dialogue topics for each domain, we leverage ChatGPT3.5 and GPT4 to generate conversations in the format of self-dialogue. The complete process of dialogue generation is illustrated in Figure~\ref{fig: dialogue generation}.

Initially, we integrate the dialogue topics into two system prompts, which are then inputted separately into two LLMs (both are ChatGPT3.5 or GPT4). These two system prompts guide the LLMs to generate $N$ rounds of dialogue in a given domain and topic. More details of the system prompts can be found in Appendix~\ref{app:system prompts}. Then, for knowledge-grounded dialogue and task-oriented dialogue, we manually examine all responses from A to determine their adherence to human language. For in both scenarios, we consider real human-machine interaction, aiming to assess the LLMs' adaptability to genuine human behaviors. (We assume that A is the user and B is the LLM. In this setup, we ensure the accuracy of A and only annotate the contents of B.) The chit-chat dialogue and reasoning dialogue are relatively unconstrained and freely conducted, without incorporating any specific human-machine interaction settings (Section 3.2). They necessitate only their memory and comprehension capabilities regarding contextual information, thereby minimizing the need for manual intervention. Consequently, for the responses of A in knowledge-grounded and task-oriented scenarios, where the contents do not conform to human language, we manually modify and have LLMs re-generate. Eventually, we obtain the complete dataset of multi-turn dialogue.

We provide one output sample and more generation details in Appendix~\ref{app:data format}. 
% highlighting the distinctive features that our DiaHalu is a naturally generated dialogue-level benchmark by LLMs, with various kinds of hallucination and explanations. 
The statistical information of the whole benchmark is in Table~\ref{tab: statistical information}.

\begin{table}[!t]
\centering
\fontsize{9pt}{9pt}\selectfont
\renewcommand{\arraystretch}{1.5} 
\begin{tabular}{cc}
\toprule
\textbf{Attribute}              & \textbf{Attribute Value} \\
\midrule
Benchmark Name                  & DiaHalu                \\
Generated by                    & ChatGPT3.5 / GPT4        \\
Sample Form                     & dialogue-level           \\
Sample Numbers                  & 1103                     \\
Dialogue Rounds                 & 6-10                     \\
Avg. Rounds                     & 6.9120                   \\
Domain Numbers                  & 4                        \\
Hallucination Subtypes             & 5                        \\
Max. Response Length (Words)    & 183                      \\
Avg. Response Length (Words)    & 13.2899                 \\
\bottomrule
\end{tabular}
\caption{The statistical information of the benchmark.}
\label{tab: statistical information}
\end{table}

\subsection{Human Annotation}
Annotating the hallucination and its types in this dataset is a very challenging task. Since there may be more than one instance of hallucination in multi-turn dialogue. Also, some hallucination subtypes in edge cases are difficult to differentiate. 
Therefore, the entire annotation process demands a high level of expertise from annotators and requires detailed definitions for ambiguous contents.

\paragraph{The annotators} of our dataset are all seasoned researchers in the field of linguistics and natural language processing. We invite experienced experts in the field of LLMs’ hallucination detection from both academia and industry to engage in discussion and conduct sampling checks. For more details about the annotators and the experts, refer to Appendix~\ref{app: supplementary for annotate} (The annotators).

\paragraph{Annotation process} is divided into three steps. (1) Each annotator labels some samples for each domain, followed by a careful discussion between the annotators and the experts. 
The discussions intricately define the application scope of each hallucination label (discussion results in Appendix~\ref{app: application scope of hallucination labels}).
% and ultimately we decide to annotate the corresponding hallucination type on the first dialogue round in which hallucination occurs. 
(2) All the annotators label the entire dataset, discussions and corrections are made for inconsistent annotations. (3) Statistical analysis is performed on the annotated results. For more details, please refer to Appendix~\ref{app: supplementary for annotate} (The Annotation Process).

\paragraph{Annotation Consistency} For evaluating the inter-annotator consistency, we calculate the Fleiss's Kappa \citep{randolph2005free} of Inter-Annotator Agreement (IAA) \citep{artstein2017inter}, which is a statistical measure used to assess the degree of agreement among multiple raters for a set of items. The final score of Fleiss's Kappa is 0.8842, representing almost perfect agreement
among all the annotators. For more calculation details, please refer to Appendix~\ref{app: supplementary for annotate} (Label Consistency).

\begin{figure}[!t]
    \centering
    \includegraphics[scale=0.5]{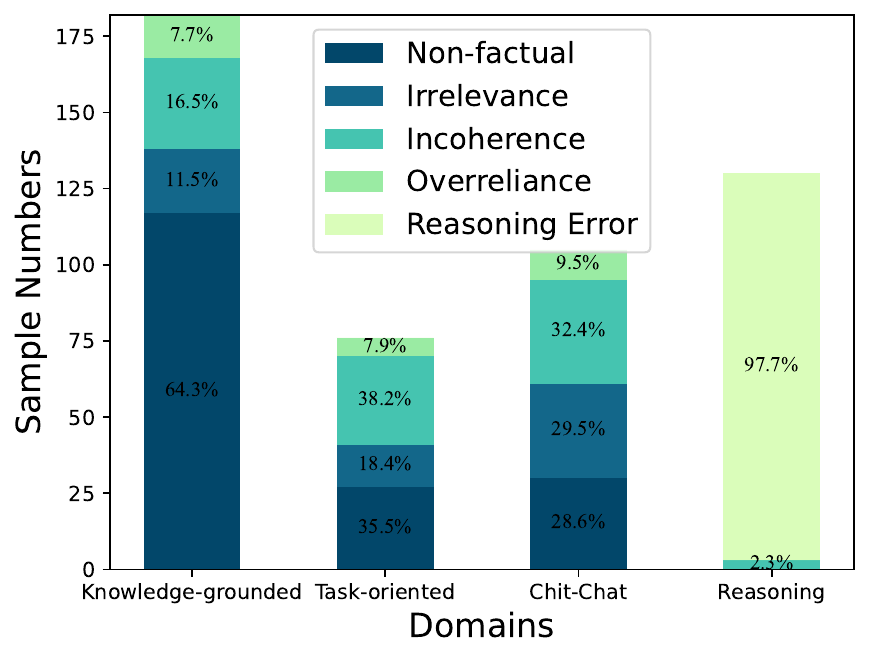}
    \caption{The distribution of five different hallucination subtypes within the four dialogue domains. }
    \label{fig: distribution of hallucination types}
\end{figure}

\paragraph{Annotation Results} After annotating the entire dataset, we conduct several statistical analyses on it. Table~\ref{tab: statistical information of hallucination} reveals the probability of hallucination occurring in each dialogue domain. The results indicate that hallucination are highly likely to arise in knowledge-grounded dialogue and reasoning dialogue. Therefore, the knowledge and reasoning abilities of LLMs still need further improvement. Despite LLMs' powerful multi-turn dialogue capability, faithfulness hallucination such as irrelevance, incoherence and overreliance still persists. Figure~\ref{fig: distribution of hallucination types} presents the proportion of each hallucination subtype in each dialogue domain. Irrelevance, incoherence, and overreliance widely exist in daily dialogue contexts, such as task-oriented and chit-chat scenarios. In knowledge-grounded dialogue, the factuality hallucination constitutes a significant proportion, while in reasoning dialogue, almost all hallucination are defined as errors in reasoning. This statistical information can help us understand the subtypes of hallucination in LLMs' multi-turn dialogue, facilitating an exploration of their origins and contributing to the elimination of these subtypes of hallucination.

\begin{table}[!t]
\centering
\fontsize{7pt}{9.6pt}\selectfont
\renewcommand{\arraystretch}{1.3} 
\begin{tabular}{p{1.5cm}p{1cm}p{0.4cm}p{0.5cm}p{1cm}p{0.7cm}}
\toprule
 & \textbf{Knowledge} & \textbf{Task} & \textbf{Chit} & \textbf{Reasoning} & \textbf{Overall}\\
 \midrule
  \midrule
 \# Number & 371 & 210 & 263 & 259 & 1103\\
\# Non-Halu & 199 & 135 & 164 & 129 & 627\\
 \# Halu & 172 & 75 & 99 & 130 & 476\\
 Halu Rate (\%) & 46.36 & 35.71 & 37.64 & 50.19 & 43.16 \\
\bottomrule
\end{tabular}
\caption{The statistical information of hallucination on the four dialogue domains. `\# number', `\# Non-Halu',`\# Halu' and `Halu Rate' represent the number of samples, the number of samples without hallucination, the number of samples with hallucination and the proportion of hallucinated samples. }
\label{tab: statistical information of hallucination}
\end{table}

\begin{table*}[!t]
\centering
\fontsize{8pt}{9pt}\selectfont
\renewcommand{\arraystretch}{1.5} 
\setlength{\tabcolsep}{1.5pt}
\begin{tabular}{c|ccc|ccc|ccc|ccc|ccc}
\toprule
\multirow{2}{*}{\textbf{Method}}  & \multicolumn{3}{|c|}{\textbf{Knowledge-grounded}}  & \multicolumn{3}{|c|}{\textbf{Task-oriented}} & \multicolumn{3}{|c|}{\textbf{Chit-Chat}} & \multicolumn{3}{|c|}{\textbf{Reasoning}} & \multicolumn{3}{|c}{\textbf{Overall}} \\
 & Precision & Recall & F1 & Precision & Recall & F1 & Precision & Recall & F1 & Precision & Recall & F1 & Precision & Recall & F1 \\
\midrule
 \midrule
Random & 41.57 & 43.02 & 42.29 & 31.86 & 48.00 & 38.30 & 38.46 & 50.51 & 43.67 & 49.61 & 49.23 & 49.42 & 40.72 & 47.06 & 43.66\\
\midrule
$SelfCheckGPT_{B}$ & 42.55 & 23.26 & 30.08 & 35.38 & 30.67 & 32.86 & 30.00 & 18.18 & 22.64 & 60.81 & 34.61 & 44.12 & 43.00 & 26.47 & 32.77\\
$SelfCheckGPT_{N}$ & 59.46 & 25.58& 35.77 & 38.84 & 62.67 & \textbf{47.96} & 45.19 & 47.47 & 46.30 & 70.58 & 18.46 & 29.27 & 48.65 & 34.03 & 40.05\\
$SelfCheckGPT_{P}$ & 55.22 & 21.51 & 30.96 & 48.00 & 32.00 & 38.40 & 45.00 & 45.45 & 45.23 & 62.37 & 44.62 & 52.02 & 52.90 & 34.45 & 41.73\\
FOCUS & 46.11 & 48.26 & \textbf{47.16} & 34.09 & 60.00 & 43.48 & 36.56 & 49.49 & 42.06 & 50.56 & 34.62 & 41.10 & 41.49 &46.64 & 43.92 \\
\midrule
LLaMa-30B & 37.50 & 5.23 & 9.18 & 30.77 & 5.33 & 9.09 & 50.00 & 11.11 & 18.18 & 81.25 & 10.00 & 17.81 & 49.33 & 7.78 & 13.43\\
Vicuna-33B & 45.45 & 5.81 & 10.31 & 42.86 & 4.00 & 7.32 & 36.36 & 4.04 & 7.27 & 51.35 & 14.62 & 22.75 & 46.75 & 7.56 & 13.02\\
\midrule
Gemini1.5 PRO &80.00 & 20.93 & 33.18 & 60.00 & 36.00 & 45.00 & 70.37 & 38.38 & \textbf{49.67} & 73.63 & 51.54 & 60.63 & 71.49 & 35.29 & 47.26 \\
ChatGPT3.5 &25.00 & 0.58 & 1.14 & 33.33 & 2.67 & 4.93 & 55.56 & 5.05 & 9.26 & 57.14 & 6.15 & 11.11 & 48.48 & 3.36 & 6.27 \\
GPT4 & 80.89 & 31.98 & 45.83 & 74.19 & 30.67 & 43.40 & 67.74 & 21.21 & 32.31 & 74.07 & 61.54 & \textbf{67.23} & 75.21 & 37.61 & \textbf{50.14} \\
\bottomrule
\end{tabular}
\caption{The classification results on four kinds of baselines, and the best F1 scores are in bold form. The indices $B$, $N$ and $P$ of SelfCheckGPT denote scoring with BERTScore, with NLI and using prompts, respectively.}
\label{tab:main result}
\end{table*}

\begin{table*}[!t]
\centering
\fontsize{8pt}{9pt}\selectfont
\renewcommand{\arraystretch}{1.5} 
\setlength{\tabcolsep}{2.5pt}
\begin{tabular}{l| lll lll lll lll | lll }
\toprule
\multirow{2}{*}{\textbf{Method}}  & \multicolumn{3}{c}{\textbf{Knowledge-grounded}}  & \multicolumn{3}{c}{\textbf{Task-oriented}} & \multicolumn{3}{c}{\textbf{Chit-Chat}} & \multicolumn{3}{c}{\textbf{Reasoning}} & \multicolumn{3}{|c}{\textbf{Overall}} \\
 & Prec & Rec & F1 & Prec & Rec & F1 & Prec & Rec & F1 & Prec & Rec & F1 & Prec & Rec & F1 \\
  \midrule
   \midrule
FaithCritic w/retrieval & 28.26 & 84.54 & 42.38 & - & - & - & - & - & - & 51.63 & 79.17 & 62.50 & - & - & -\\
\midrule
Gemini1.5 PRO &80.00 & 20.93 & 33.18 & 60.00 & 36.00 & 45.00 & 70.37 & 38.38 & 49.67 & 73.63 & 51.54 & 60.63 & 71.49 & 35.29 & 47.26 \\
w/ CoT & 81.25  & 22.67 & 35.45{\color{blue}$\uparrow$}& 69.77 & 40.00 & 50.85{\color{blue}$\uparrow$}& 75.00 & 36.36 & 48.98{\color{red}$\downarrow$} & 72.92 & 53.85 & 61.95{\color{blue}$\uparrow$} & 74.47 & 36.76 & 49.23{\color{blue}$\uparrow$} \\
w/ one-shot & 80.43 & 21.51 & 33.94{\color{blue}$\uparrow$} & 60.87 & 37.33 & 46.28{\color{blue}$\uparrow$} & 70.91 & 39.39& 50.65{\color{blue}$\uparrow$} & 73.91 & 52.31 & 61.26{\color{blue}$\uparrow$} & 71.97 & 36.13 & 48.11{\color{blue}$\uparrow$} \\
w/ retrieval & 86.04 & 21.51 & 34.42{\color{blue}$\uparrow$} & - & - & - & - & -& -& 76.70 & 60.77 & 67.81{\color{blue}$\uparrow$} & - & - & - \\
\midrule
ChatGPT3.5 &25.00 & 0.58 & 1.14 & 33.33 & 2.67 & 4.94 & 55.56 & 5.05 & 9.26 & 57.14 & 6.15 & 11.11 & 48.48 & 3.36 & 6.27 \\
w/ CoT & 45.45  & 2.91 & 5.46{\color{blue}$\uparrow$} & 33.33 & 2.67 & 4.94 & 40.00 & 4.04 & 7.34{\color{red}$\downarrow$} & 47.06 & 6.15 & 10.88{\color{blue}$\uparrow$} & 43.18 & 3.99 & 7.31{\color{blue}$\uparrow$} \\
w/ one-shot & 40.00 & 1.16 & 2.26{\color{blue}$\uparrow$} & 42.85 & 4.00 & 7.31{\color{blue}$\uparrow$} & 46.15 & 6.06& 10.71{\color{blue}$\uparrow$} & 56.25 & 6.92 & 12.32{\color{blue}$\uparrow$} & 48.78 & 4.20 & 7.73{\color{blue}$\uparrow$} \\
w/ retrieval & 70.00 & 4.01 & 7.69{\color{blue}$\uparrow$} & - & - & - & - & -& -& 70.58 & 9.23 & 16.32{\color{blue}$\uparrow$} & - & - & - \\
 \midrule
GPT4 & 80.89 & 31.98 & 45.83 & 74.19 & 30.67 & 43.40 & 67.74 & 21.21 & 32.31 & 74.07 & 61.54 & 67.23 & 75.21 & 37.61 & 50.14 \\
w/ CoT & 86.05 & 21.51 & 34.42{\color{red}$\downarrow$} & 73.17 & 40.00 & 51.72{\color{blue}$\uparrow$} & 80.56 & 29.29 & 42.96{\color{blue}$\uparrow$} & 71.43 & 76.92 & 74.07{\color{blue}$\uparrow$} & 75.38 & 41.18 & 53.26{\color{blue}$\uparrow$}\\
w/ one-shot & 81.42 & 33.14 & 47.11{\color{blue}$\uparrow$} & 71.87 & 30.67 & 42.99{\color{red}$\downarrow$} & 72.22 & 26.26& 38.52{\color{blue}$\uparrow$} & 73.11 & 66.92 & 69.87{\color{blue}$\uparrow$} & 75.09 & 40.55 & 52.66{\color{blue}$\uparrow$} \\
w/ retrieval & 77.89 & 43.02 & 55.43{\color{blue}$\uparrow$} & - & - & - & - & -& -& 74.40 & 71.54 & 72.94{\color{blue}$\uparrow$} & - & - & - \\
\bottomrule
\end{tabular}
\caption{The results of CoT, the one-shot settings and the retrieval technique on the three closed-source LLMs. The {\color{blue}$\uparrow$} and {\color{red}$\downarrow$} indicate whether these can promote improvements in F1 score. We also provide the detection results of FaithCritic with the retrieval technique.}
\label{tab: CoT and retrieval}
\end{table*}

% \begin{table*}[!t]
% \centering
% \fontsize{8pt}{9pt}\selectfont
% \renewcommand{\arraystretch}{1.5} 
% \setlength{\tabcolsep}{2.1pt}
% \begin{tabular}{c|ccc|ccc|ccc|ccc|ccc|}
% \toprule
% \multirow{2}{*}{\textbf{Method}}  & \multicolumn{3}{|c|}{\textbf{Knowledge-grounded}}  & \multicolumn{3}{|c|}{\textbf{Task-oriented}} & \multicolumn{3}{|c|}{\textbf{Chit-Chat}} & \multicolumn{3}{|c|}{\textbf{Reasoning}} & \multicolumn{3}{|c|}{\textbf{Overall}} \\
%  & Precision & Recall & F1 & Precision & Recall & F1 & Precision & Recall & F1 & Precision & Recall & F1 & Precision & Recall & F1 \\
%   \midrule
%    \midrule
% checkall_{3.5} &25.00 & 0.58 & 1.14 & 33.33 & 2.67 & 4.93 & 55.56 & 5.05 & 9.26 & 57.14 & 6.15 & 11.11 & 48.48 & 3.36 & 6.27 \\
% selfcheck_{3.5} & 50.00  & 0.76 & 1.50 & 0.00 & 0.00 & 0.00 & 42.86 & 4.23 & 7.69 & 55.56 & 6.25 & 11.24 & 45.00 & 2.74 & 5.16 \\
%  \midrule
% checkall_{3.5} & 80.89 & 31.98 & 45.83 & 74.19 & 30.67 & 43.40 & 67.74 & 21.21 & 32.31 & 74.07 & 61.54 & 67.23 & 75.21 & 37.61 & 50.14 \\
% selfcheck_{4} & 72.72 & 19.51 & 30.77 & 75.00 & 32.14 & 45.00 & 100.00 & 28.57 & 44.44 & 88.24 & 30.00 & 44.78 & 83.33 & 27.21 & 41.02\\
% \bottomrule
% \end{tabular}
% \caption{SelfCheck.}
% \label{tab: selfcheck}
% \end{table*}

\section{Experiments}
In this section, we assess the performance of several evaluation models and specialized methods on the dataset we introduced. 
Thereby, we can trial the effectiveness of existing methods in detecting dialogue-level hallucination.
We still conduct more fine-grained detection and explore whether the phenomenon of hallucination snowballing exists.

\subsection{Baselines}
We select some powerful LLMs to detect hallucination by providing specific prompts. These models include open-source LLMs: LLaMa-30B \citep{DBLP:journals/corr/abs-2302-13971}, Vicuna-33B \citep{23vicuna}, and some closed-source LLMs: Gemini1.5 PRO \citep{DBLP:journals/corr/abs-2312-11805}, ChatGPT3.5 \citep{10113601} and GPT4 \citep{openai2023gpt4}. Similarly, we also experiment on specialized existing hallucination detection methods, such as FaithCritic \citep{DBLP:journals/tacl/DziriKMZYPR22}, SelfCheckGPT \citep{DBLP:conf/emnlp/ManakulLG23} and FOCUS \citep{DBLP:conf/emnlp/ZhangQGDZZZWF23}. For a detailed description of the above baselines, please refer to Appendix~\ref{app: baseline} (I. Baselines Selected).

\subsection{Metrics}
For hallucination detection, we use standard binary classification to determine whether there exists hallucination (Table~\ref{tab:main result}). We utilize binary classification evaluation metrics: Precission, Recall and F1. The
positive label for this classification task is set as "Halu". Meanwhile, we also conduct more fine-grained hallucination-type recognition to judge the specific subtype of hallucination and use micro-F1 score for all hallucination categories (Table~\ref{tab: xilidu}). Appendix~\ref{app: baseline} (II. Metrics Calculation) provides more thorough explanations.

\subsection{Main Results}
From the results in Table~\ref{tab:main result}, we can get the following conclusions. 

\paragraph{First, DiaHalu is a highly challenging benchmark for dialogue-level hallucination detection.} 
Except for GPT4, the F1 scores of all other detection methods and detecting LLMs do not exceed 50.00. 
Existing LLMs, such as LLaMa-30B and Vicuna-33B, are not effective in accurately discerning most samples that involve hallucination.
Regarding the specialized detection methods FOCUS and SelfCheckGPT (applying prompt and NLI methods), they achieve relatively better performances. However, it proves challenging with BERTScore for SelfCheckGPT.
% However, distinguishing between positive and negative samples proves challenging with BERTScore for SelfCheckGPT.

\paragraph{Second, ChatGPT3.5 shows a noticeable phenomenon of overconfidence.} Our dataset is primarily generated by ChatGPT3.5, which exhibits high confidence in its output. Despite providing a specially designed detection prompt, it still struggles to differentiate whether the dialogue content is hallucinated or not, not along the samples generated by GPT4. So, the majority of its output labels are "Non-Halu".

\paragraph{Third, the faithfulness hallucination is more difficult to detect for LLMs.} Apart from the specialized hallucination detection methods, 
the results from directly harnessing LLMs for judgment indicate that the recognition accuracy for task-oriented and chit-chat domains of dialogue are much lower than that for the knowledge-grounded and reasoning dialogue. 
This is because the hallucination types in the knowledge-grounded and reasoning dialogue are primarily Non-factual and Reasoning Error, which present in a more direct and apparent manner.
Nevertheless, task-oriented and chit-chat domains mainly consist of three subtypes of faithfulness hallucination, which requires a LLM to possess long-term memory and the ability to recognize topics/roles transition in dialogue.

\begin{table}[!t]
\fontsize{7pt}{11pt}\selectfont
\begin{tabular}{c|ccccc|c}
\toprule
              & \textbf{NF}    & \textbf{Ic}    & \textbf{Ir}    & \textbf{Ov}   & \textbf{RE}    & \textbf{ALL}   \\ 
\midrule
Gemini1.5 PRO & 18.97 & 30.49 & 11.36 & 4.76 & 45.41 & 26.72 \\ 
w/ one-shot   & \textbf{19.74} & \textbf{32.53} & \textbf{11.49} & 4.65 & \textbf{46.24} & \textbf{27.96} \\ 
\midrule
ChatGPT3.5    & 1.16  & 4.26  & 0.00  & 0.00 & 9.66  & 3.93  \\ 
w/ one-shot   & \textbf{1.18}  & \textbf{4.35}  & 0.00  & \textbf{6.45} & \textbf{10.96} & \textbf{4.64} \\ 
\midrule
GPT4          & 29.38 & 25.00 & 5.71  & 4.65 & 55.66 & 32.30 \\ 
w/ one-shot   & \textbf{31.34} & 21.95 & \textbf{5.41}  & \textbf{8.33} & \textbf{60.00} & \textbf{34.11} \\ 
\bottomrule
\end{tabular}
\caption{Fine-grained hallucination-type recognition F1 scores for three LLMs. `NF', `Ic', `Ir', `Ov' and `RE' stand for Non-factual, Incoherence, Irrelevance, Overreliance and Reasoning Error, respectively. `ALL' represents micro-f1 of all hallucination subtypes. We also incorporate a sample in the prompt to assist with the judgment. If the results in the one-shot setting show improvement, we bold them.}
\label{tab: xilidu}
\end{table}

\begin{figure}[!t]
    % \centering
    \includegraphics[scale=0.3]{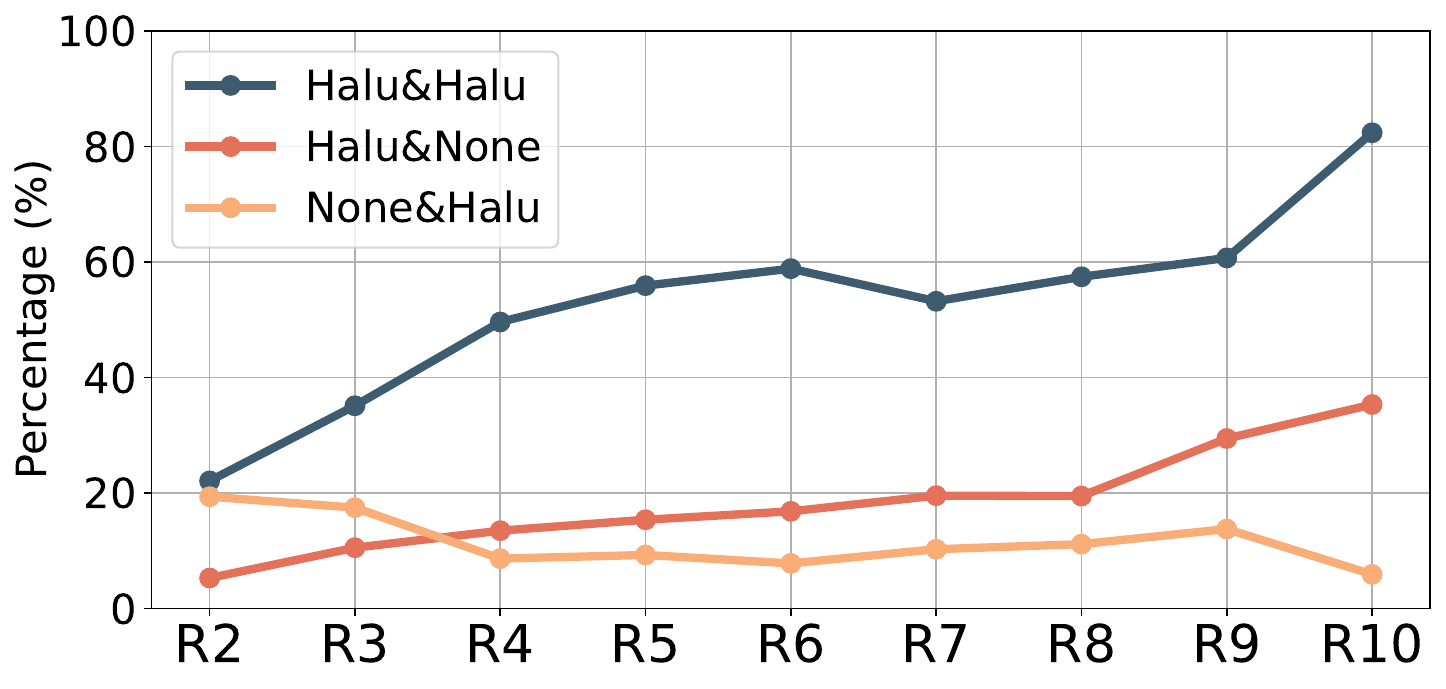}
    \caption{The proportions of the three dialogue round categories. For example, the three values of R7 denote the proportions of `these three categories in the 7th round' within `hallucinated samples that have at least seven rounds dialogues'.}
    \label{fig: rounds}
\end{figure}

\subsection{Chain-of-Thought, One-shot and Retrieval for Detection}
Chain-of-Thought (CoT) and Retrieval are two important techniques for enhancing the ability of LLMs.
In this section, we test whether these two techniques can improve the effectiveness of hallucination detection in Table~\ref{tab: CoT and retrieval}. More details are depicted in Appendix~\ref{app: CoT and Retrieval}. Meanwhile, we also incorporate a sample based into the vanilla prompt to test the few-shot capabilities of these three language models.

The experimental results indicate that all these three methods have facilitating effects on hallucination detection. However, Gemini1.5 PRO and ChatGPT3.5 with CoT show a decrease of around 1.00 F1 points in the chit-chat domain. We believe that these two models inherently lack the ability to recognize faithfulness hallucination, and additional CoT contents introduce noise to their judgments.

\subsection{Fine-grained Hallucination-type Recognition}
Table~\ref{tab: xilidu} shows fine-grained hallucination-type recognition results for three open-source LLMs. We can conclude that ChatGPT3.5 fails the recognition of almost all labels. To some extent, Gemini1.5 PRO and GPT4 have the ability to recognize factuality hallucination and reasoning errors, but they have lower F1 scores for the three subtypes of faithfulness hallucination. This result reveals that faithfulness hallucination remains a pressing issue for LLMs. In the one-shot setting, all three models show improved accuracy in recognizing most labels. However, this setting also introduces some sample noise that interferes with the models' judgments, such as GPT4 decreases its recognition F1 of the `Ic' label.

\subsection{Hallucination Snowballing}
In this section, we study the hallucination snowballing phenomenon \citep{DBLP:journals/corr/abs-2309-01219} in our benchmark. Specifically, for each round of dialog (2-10) in all hallucinated samples, we define three categories: \textbf{I} hallucination that appears in previous rounds and also appears in the current round (Halu\&Halu), \textbf{II} hallucination that appears in previous rounds but not appear in the current round (Halu\&None) and \textbf{III} hallucination that not appear in previous rounds but appears in the current round (None\&Halu). We calculate the proportions of these three categories in Figure~\ref{fig: rounds}.

First, \textbf{I} is greater than the other two categories (\textbf{II} and \textbf{III}), which means that hallucinated contents are more likely to generate new hallucinated responses. Second, \textbf{I} shows the most obvious increasing trend, indicating that the probability of hallucination increases with the number of dialogue rounds. These two findings validate the hallucination snowballing phenomenon.

\section{Conclusion}
In this paper, we propose a dialogue-level hallucination evaluation benchmark named DiaHalu. We construct the benchmark in a three-step process. The DiaHalu covers four multi-turn dialogue domains and five hallucination subtypes. Experiments through some well-known LLMs and specialized detection methods on the benchmark show that it is a challenging task, holding significant value for further research (Appendix~\ref{app: future works}).

\section*{Limitations}
This paper proposes a novel dedicated dialogue-level hallucination detection evaluation benchmark named DiaHalu. The benchmark covers four multi-turn dialogue domains and five hallucination subtypes. There is significant value for further research. However, two main limitations also exist. 
(1) During the second step of the benchmark construction phase, aligning the contents of speaker A with human language consumes a considerable amount of time and effort. 
Frequent calls to the ChatGPT3.5 or GPT4 API Keys result in a significant expense. 
Simultaneously, achieving consistency among all annotators led to prolonged discussion time and money cost.
(2) 
We do not partition the dataset into training, validation, and test sets. The primary purpose of evaluation benchmarks is to assess a models' capabilities. However, if we divide the dataset into the above three categories, this is about assigning capabilities to models. From the perspective of the two objectives, there is a clear difference. Another reason is that we need to consider the black-box detection scenario for those closed-source LLMs. However, if a division into these three types of datasets is necessarily required, it would require more data samples and larger resource consumption. 

\section*{Ethics Statement}
The benchmark is primarily generated by ChatGPT3.5 or GPT4. We obtain all the API Keys through a paid subscription. All the annotators are real people and they receive corresponding compensation and rewards. The entire process and outcomes are free from intellectual property and ethical legal disputes.

\section*{Acknowledgements}
This research is funded by the National Science and Technology Major Project (No. 2021ZD0114002), the National Nature Science Foundation of China (No. 62477010, No. 62307028), the Science and Technology Commission of Shanghai Municipality Grant (No. 22511105901, No. 21511100402), and the Shanghai Science and Technology Innovation Action Plan (No. 23ZR1441800 and No. 23YF1426100).

We would like to thank all the reviewers for providing suggestions to improve our work.

% Firstly, we would like to thank all the authors and reviewers for their assistance in our work. The idea for this paper comes from my old friend Hill Zhang in ByteDance, we realize the importance of dialog-level hallucination during our casual conversations about the application of agents. Secondly, the structure of this paper is inspired by the work FELM of Shiqi Chen from the City University of Hong Kong. Lastly, my roommate Yinqi Zhang generously provides me with a GPT4 account for completing some of the experiments.

\bibliography{acl_latex}

\appendix

\section{Appendices}

\subsection{The Comparison with Other Benchmarks}
\label{app: comparison}
In Table~\ref{tab: comparison}, we present the differences between DiaHalu and other hallucination detection benchmarks, highlighting the distinctive features that our DiaHalu is a 
naturally generated dialogue-level benchmark by LLMs, with various kinds of hallucination and explanations. All the compared benchmarks can be referred to in Section 2.1. HaluEval, WikiBio+, and PHD benchmark come from the paper \citet{DBLP:conf/emnlp/LiCZNW23}, \citet{DBLP:conf/emnlp/ManakulLG23} and \citet{DBLP:conf/emnlp/YangS023} respectively.

\subsection{The Four Dialogue Domains}
\label{app:four domains}

\paragraph{Knowledge-grounded dialogue} is designed for users to engage in knowledge-based dialogue with LLMs \citep{DBLP:conf/aaai/GhazvininejadBC18}.
The knowledge includes world knowledge, factual knowledge, commonsense knowledge, and multi-hop web knowledge.
It principally examines the accuracy of knowledge of the parameters in LLMs \citep{DBLP:conf/emnlp/PetroniRRLBWM19}.

\paragraph{Task-oriented dialogue} is in a form of human-computer interaction, intending to accomplish a user-specified task \citep{DBLP:conf/eacl/Rojas-BarahonaG17}. 
This type of dialogue system focuses on understanding the users' task requirements and utilizes a LLM to provide relevant information or perform specific tasks accordingly.

\paragraph{Chit-Chat dialogue} involves open-ended and non-goal dialogue \citep{DBLP:conf/naacl/SunMCRSLWLCC21}. 
We provide two LLMs with personas and facilitate a dialogue between them. 
This approach allows for the evaluation of their memory capabilities, conversational coherence, and relevance to the information being discussed.

\paragraph{Reasoning dialogue} centralizes on the logical reasoning and understanding capabilities of LLMs. Following previous works \citep{DBLP:journals/corr/abs-2310-00741,DBLP:journals/corr/abs-2311-09702,DBLP:journals/corr/abs-2401-00290,DBLP:journals/corr/abs-2402-19405,zheng2023does,DBLP:conf/acl/0009C23}, we also treat reasoning errors as a kind of hallucination.
We have the models discuss mathematical problems to achieve the answers \citep{DBLP:journals/corr/abs-2401-03238}.

\subsection{The Distribution of the Sources for Dialogue Topics}
\label{app: distribution}
\begin{figure}[!t]
    \centering
    \includegraphics[scale=0.3]{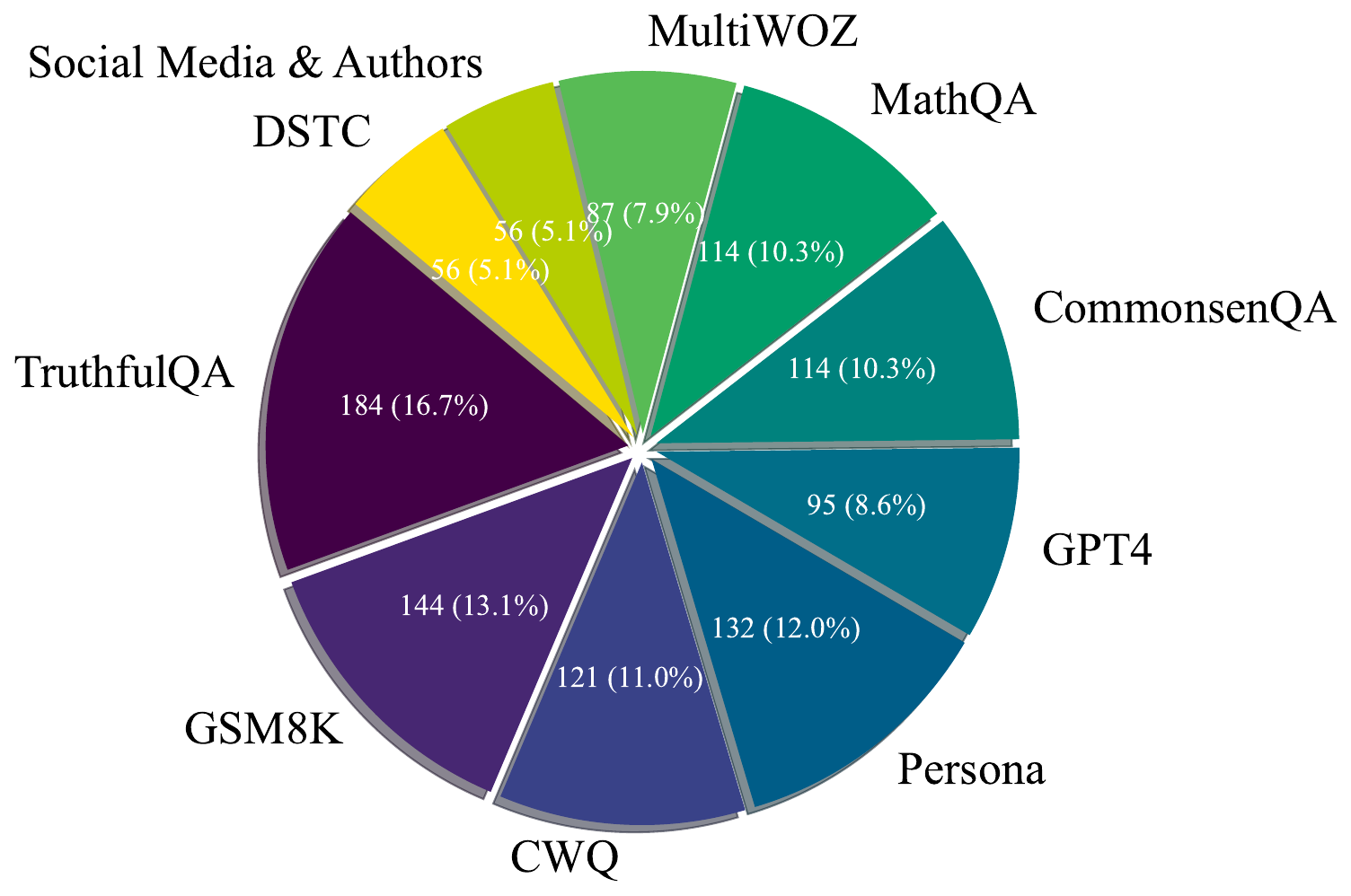}
    \caption{The distribution of the 10 sources for dialogue topics.}
    \label{fig: distribution}
\end{figure}

The number and proportion of all dialogue topics in DiaHalu across the 10 topic sources are shown in Figure~\ref{fig: distribution}. We use the hot topics of the social media, such as Facebook and Twitter, to obtain more topics in data generation process. GPT4 is used to augment and elaborate on the topics, scenes, and user behaviors. Take MultiWOZ dataset (which covers 7 real-life scenarios) as an example, we generate more real-life task-oriented dialogue scenarios via GPT4. As for the DSTC dataset with a focus on bus routes, we apply GPT4 to generate more potential user behaviors related to buses.

\subsection{System Prompts for Dialogue Generation}
\label{app:system prompts}
\begin{figure*}[!t]
    \centering
    \includegraphics[scale=0.5]{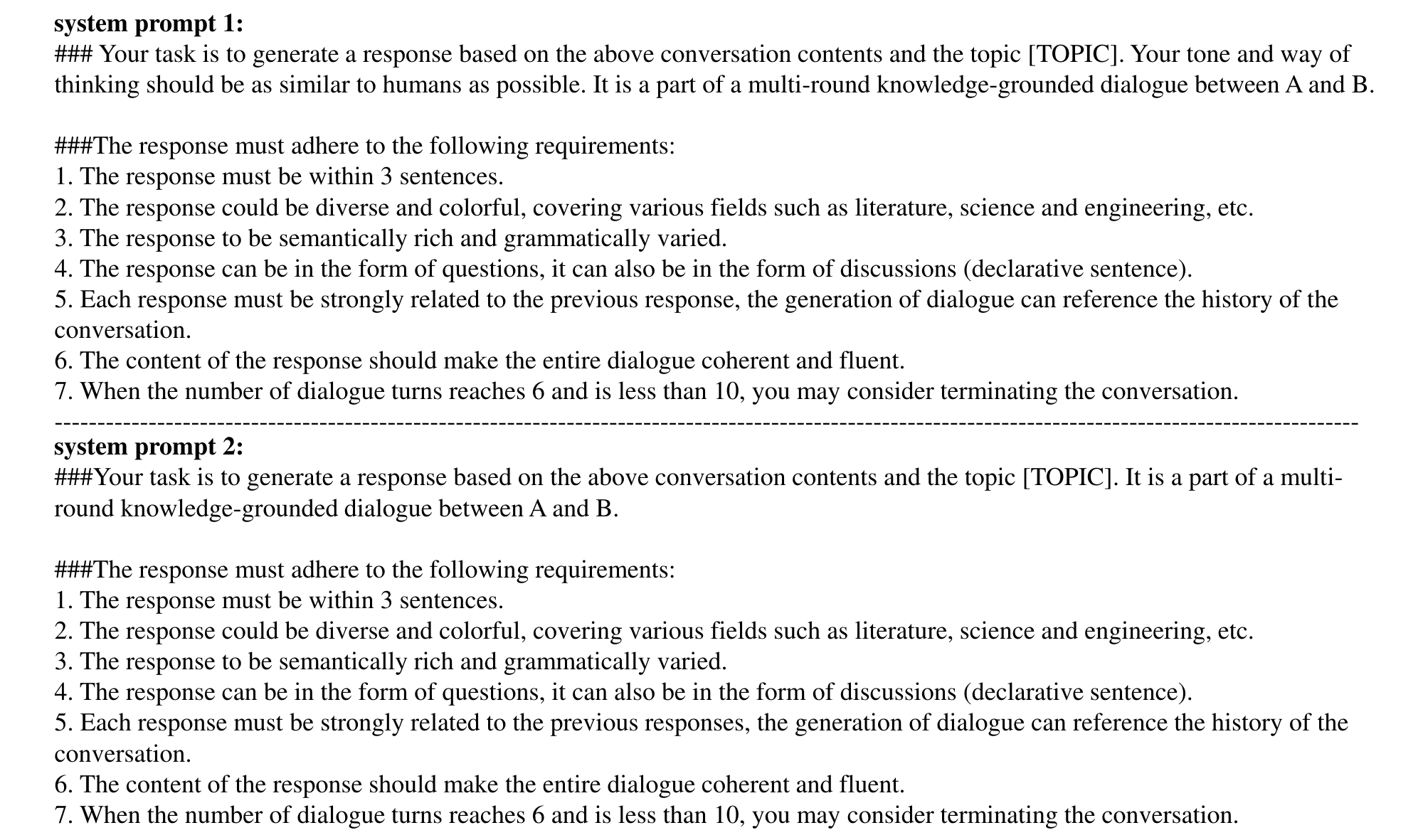}
    \caption{The brief system prompts for knowledge-grounded dialogue.}
    \label{fig:system prompt for knowledge-grouded}
\end{figure*}

\begin{figure*}[!t]
    \centering
    \includegraphics[scale=0.5]{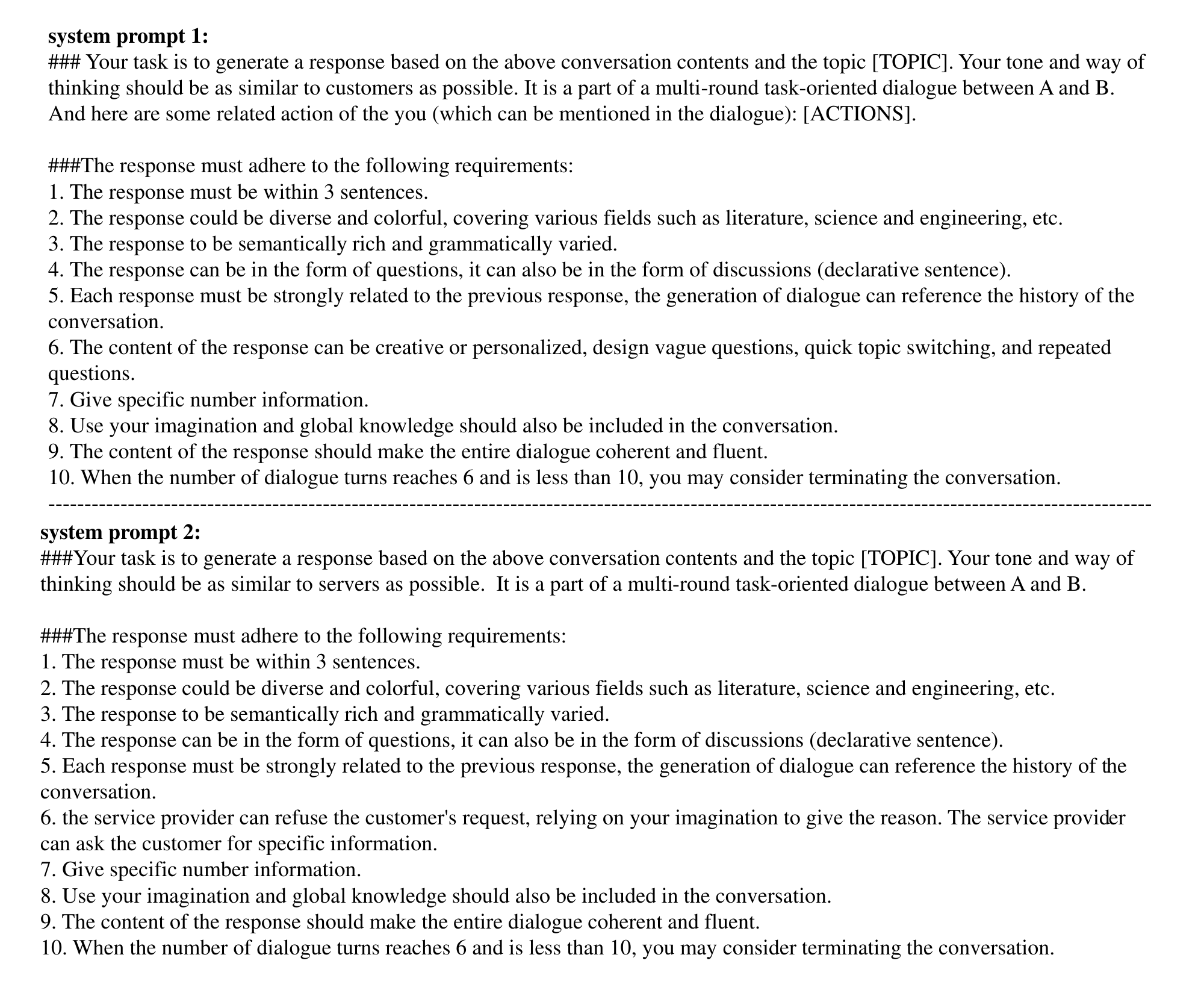}
    \caption{The brief system prompts for task-oriented dialogue.}
    \label{fig:system prompt for task-oriented}
\end{figure*}

\begin{figure*}[!t]
    \centering
    \includegraphics[scale=0.5]{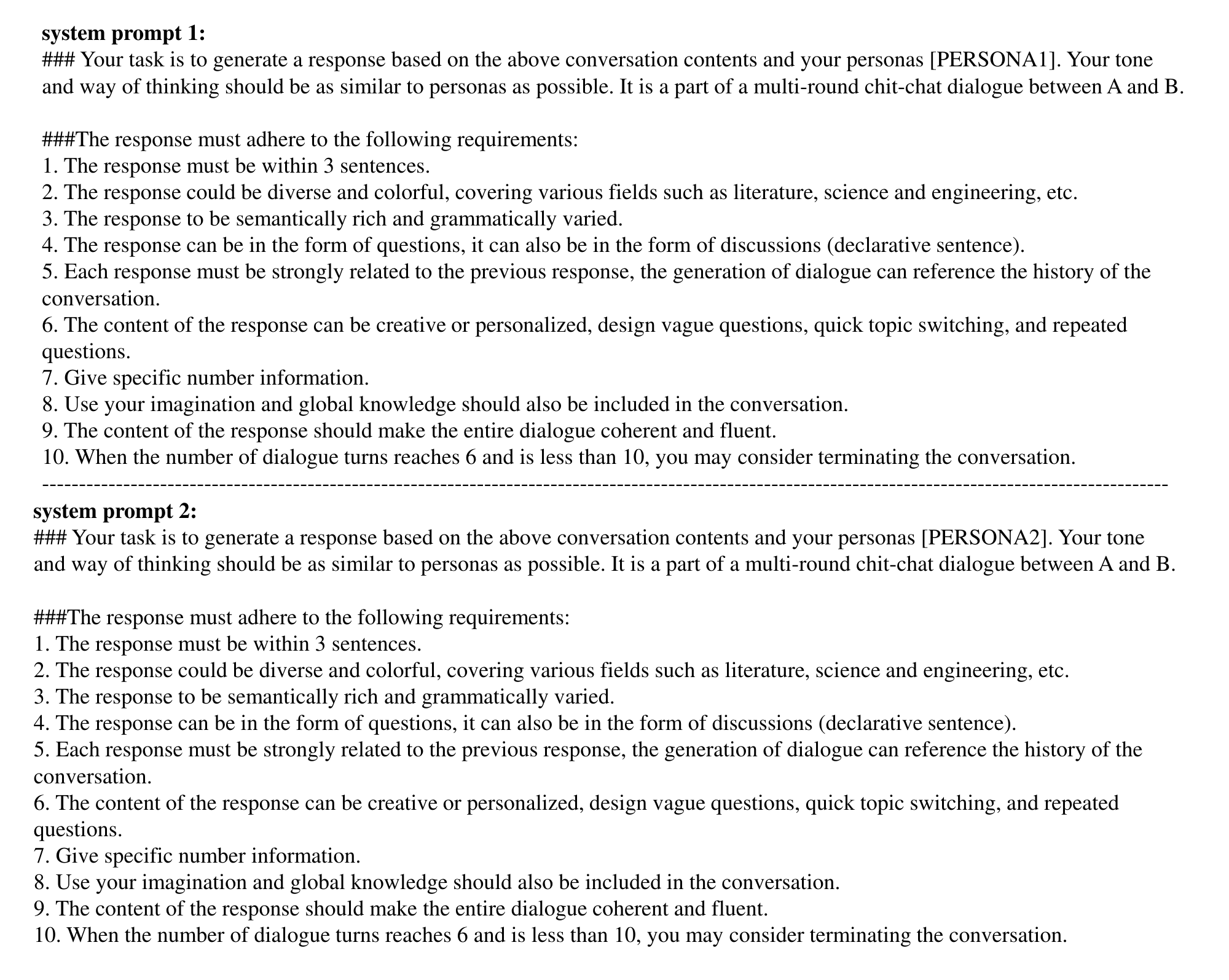}
    \caption{The brief system prompts for chit-chat dialogue.}
    \label{fig:system prompt for chit-chat}
\end{figure*}

\begin{figure*}[!t]
    \centering
    \includegraphics[scale=0.5]{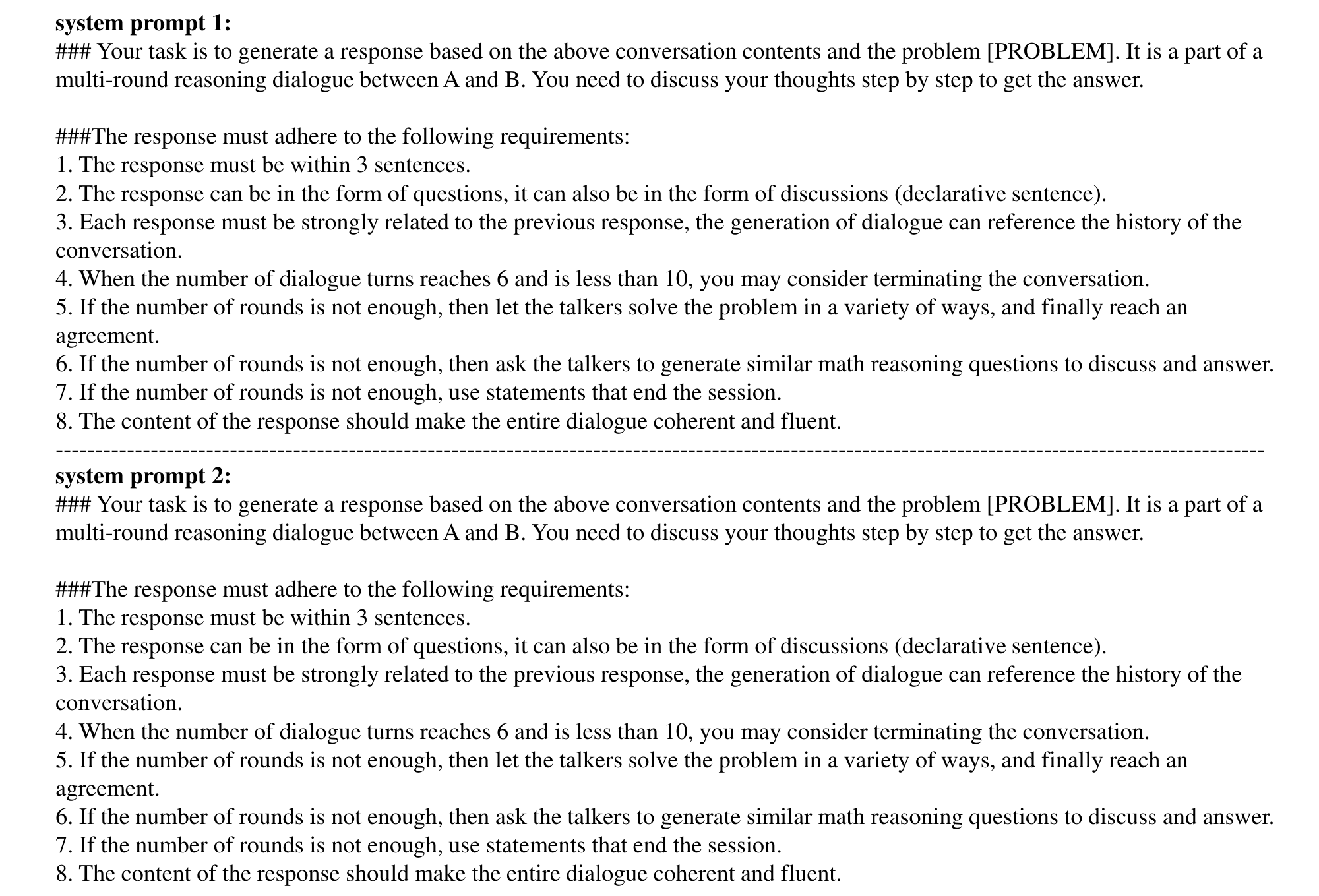}
    \caption{The brief system prompts for reasoning dialogue.}
    \label{fig:system prompt for reasoning}
\end{figure*}

In this section, we present the specific form of the system prompts for the four dialogue domains. We use ChatGPT3.5 (gpt-3.5-turbo-1106)\footnote{https://platform.openai.com/docs/models/gpt-3-5-turbo} and GPT4 (gpt-4-1106-preview)\footnote{https://openai.com/gpt-4} with the temperature 0.1. The brief system prompts of the knowledge-grounded dialogue, task-oriented dialogue, chit-chat dialogue and reasoning dialogue are respectively presented in Figure~\ref{fig:system prompt for knowledge-grouded}, Figure~\ref{fig:system prompt for task-oriented}, Figure~\ref{fig:system prompt for chit-chat} and Figure~\ref{fig:system prompt for reasoning}.

Previous works prove the LLMs' ability to follow complex instructions \citep{DBLP:conf/emnlp/ManakulLG23,DBLP:journals/corr/abs-2310-14564,DBLP:journals/corr/abs-2310-00741,mündler2023selfcontradictory,DBLP:conf/emnlp/LiCZNW23,DBLP:journals/corr/abs-2303-04048,DBLP:conf/emnlp/LiuIXWXZ23,DBLP:journals/corr/abs-2401-07103}, including some hallucination tasks. Thus, we reference such kinds of prompts, and then we formulate prompts for our benchmark.

\subsection{Generation Details and Data Format}
\label{app:data format}
We utilize 8 ChatGPT3.5 and 2 GPT4 API Keys, generating approximately 4000 examples in total initially. We conduct a thorough examination of samples to filter the confused formats or contents not aligned with instructions (system prompts). To ensure a balanced number of positive and negative samples, we manually remove samples with highly similar semantics. Finally, the number of samples generated by ChatGPT3.5 and GPT4 are 748 and 355 respectively. The total number is 1103. The reason we use GPT4 to generate data is to ensure its competitiveness for a long time in the future. 

Previous well-known hallucination benchmarks \citep{DBLP:conf/emnlp/ManakulLG23,DBLP:journals/corr/abs-2310-14564,DBLP:journals/corr/abs-2310-00741,DBLP:conf/emnlp/YangS023} contain 238, 400, 847, and 300 samples respectively, which is at the similar scale as ours. What's more, our benchmark is at dialogue level, which contains about 7620 rounds (more than 27600 rounds initially) of interactions in total (6.9120 average rounds per dialogue as shown in Table 2). This indicates a larger volume of data compared to previous benchmarks at the sentence and passage levels.

% According to the authors' estimation, our 1,103 generated dialogue samples are selected from 4000 generated dialogue samples in total.

It is also worth noting that, given that we assume the two subjects of the dialogue are A and B, both A and B are set to be either ChatGPT3.5 or GPT4, and it is not possible for one to be ChatGPT3.5 and the other to be GPT4.

Humans usually respond with flexible and brief expressions in dialogues, so we remove some GPT-specific generation patterns to better simulate natural human language. These patterns include special phrases like "I am an AI, so I cannot answer the corresponding question", detailed sectional explanations of a concept or method, excessively long sentences, and so on. Chit-Chat and Reasoning dialogue can also represent human-computer interaction. According to our preliminary studies, the formatting errors and GPT-specific generation patterns (mentioned above) are less likely to occur in these two types of dialogues, thus we skip the manual examination of whether A’s response adheres to human language, and focus on other examinations such as topic diversity. 

\begin{figure*}[!t]
    \centering
    \includegraphics[scale=0.5]{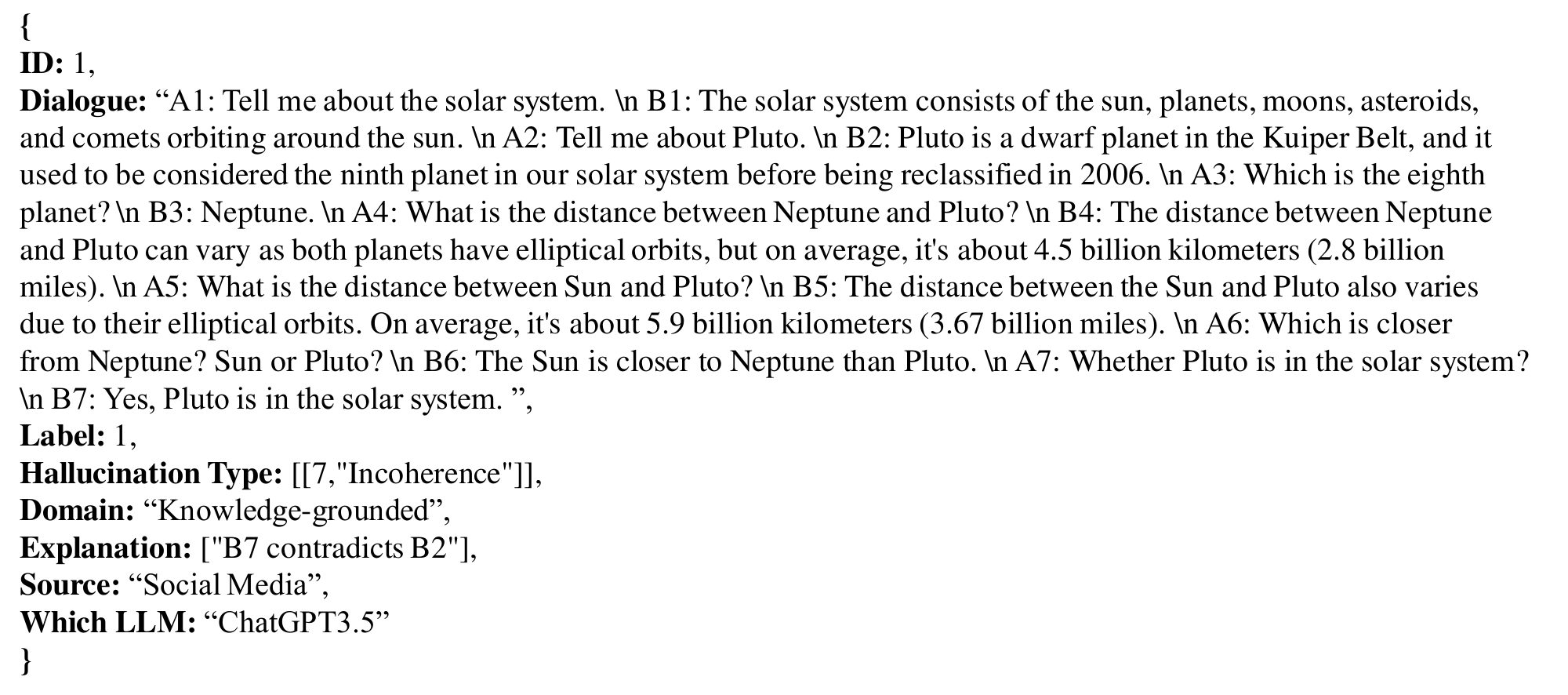}
    \caption{One sample from our benchmark DiaHalu}
    \label{fig: one sample from the benchmark}
\end{figure*}

We provide the specific format of one sample from the benchmark in Figure~\ref{fig: one sample from the benchmark}. 

\subsection{The Supplementary Details for Annotation}
\label{app: supplementary for annotate}
\paragraph{The annotators} all obtain at least a bachelor's degree, get a high score in IELTS or TOEFL exams, and are proficient in using search engines such as Google and Bing. The annotators are all seasoned researchers in the field of linguistics and natural language processing. In addition, everyone exhibits strong collaborative and communicative skills. We also invite senior experts in the field of hallucination detection from academia and industry to engage in discussions and data checking.

\paragraph{The Annotation Process} To ensure the annotation quality, we perform three steps for annotation as described in Section 4.3 (Annotation process). \textbf{First}, each annotator labels around 50 samples for each domain. The annotators are required to label the presence of hallucination, hallucination subtypes and locations, along with the corresponding explanations. For cases of inconsistent annotation, we invite experts to provide suggestions in a discussion. After that, annotators specify the application scope of each hallucination label as needed. Then the annotators take a vote for resolving the label-inconsistency of the first 50 samples in each domain. \textbf{Second}, the entire dataset is annotated according to this standard. The annotators label all the rest samples and vote for the inconsistent samples, following data checks by the experts. 
\textbf{Third}, we conduct data statistics of the whole dataset.

In the first step above, the discussion is organized in the form of online meetings. Annotators provide the inconsistent-labeled samples to experts (first 50 in each domain), after which all annotators and experts agree on a time for an online meeting discussion. Experts provide suggestions, and annotators modify the application scopes of hallucination labels based on the suggestions, thus making it more reliable.

\paragraph{The Price} The annotation time for each sample ranges from 2 to 10 (average 6.2) minutes. We pay each annotator 0.5 US dollars for annotating a sample and pay each expert 0.5 US dollars for checking a sample. This exceeds the local average hourly wage. Through the aforementioned approach, the quality of the annotations and the value of the benchmark are ensured. Thus, we consider it is greatly contributory to propose such a benchmark.

\paragraph{Label Consistency} After the whole annotation process, we achieve a label matrix $\mathbf{L} \in \mathbb{R}^{N_s*N_A}$. $N_s, N_A$ represent the number of dialogue samples and annotators respectively. The calculation for Fleiss’s Kappa is shown below:
\begin{equation}
\begin{split}
    P_e &=  \left( \frac{\sum_{i=1}^{N_s} \sum_{j=1}^{N_A} \chi_{\left\{0\right\}}(\mathbf{L}\left[i,j\right])}{N_s*N_A} \right)^2  \\
    &+ \left( \frac{\sum_{i=1}^{N_s} \sum_{j=1}^{N_A} \chi_{\left\{1\right\}}(\mathbf{L}\left[i,j\right])}{N_s*N_A} \right)^2
\end{split}
\end{equation}
\begin{equation}
\begin{split}
    P_o &= \frac{1}{N_s}  \sum_{i=1}^{N_s}   \frac{\left(\sum_j^{N_A}\chi_{\left\{0\right\}}(\mathbf{L}\left[i,j\right])\right)^2}{{N_A*\left(N_A-1\right)}}  \\ 
    &+ \frac{\left(\sum_j^{N_A}\chi_{\left\{1\right\}}(\mathbf{L}\left[i,j\right])\right)^2 -N_A}{N_A*\left(N_A-1\right)} 
\end{split}
\end{equation}
\begin{equation}
    Fleiss's Kappa = \frac{P_o-P_e}{1-P_e}
\end{equation}
where $P_o$ and $P_e$ represent the relative observed agreement among annotators and the hypothetical probability of chance agreement respectively. $\chi_{y}(Y)$ is the Indicator Function, which means when the value of $Y$ is in set $y$, the whole function equals 1.

According to the above formulas, the calculated result for Fleiss’s Kappa of our benchmark is 0.8842, representing almost perfect agreement among all the annotators.

\begin{figure*}[!t]
    \centering
    \includegraphics[scale=0.5]{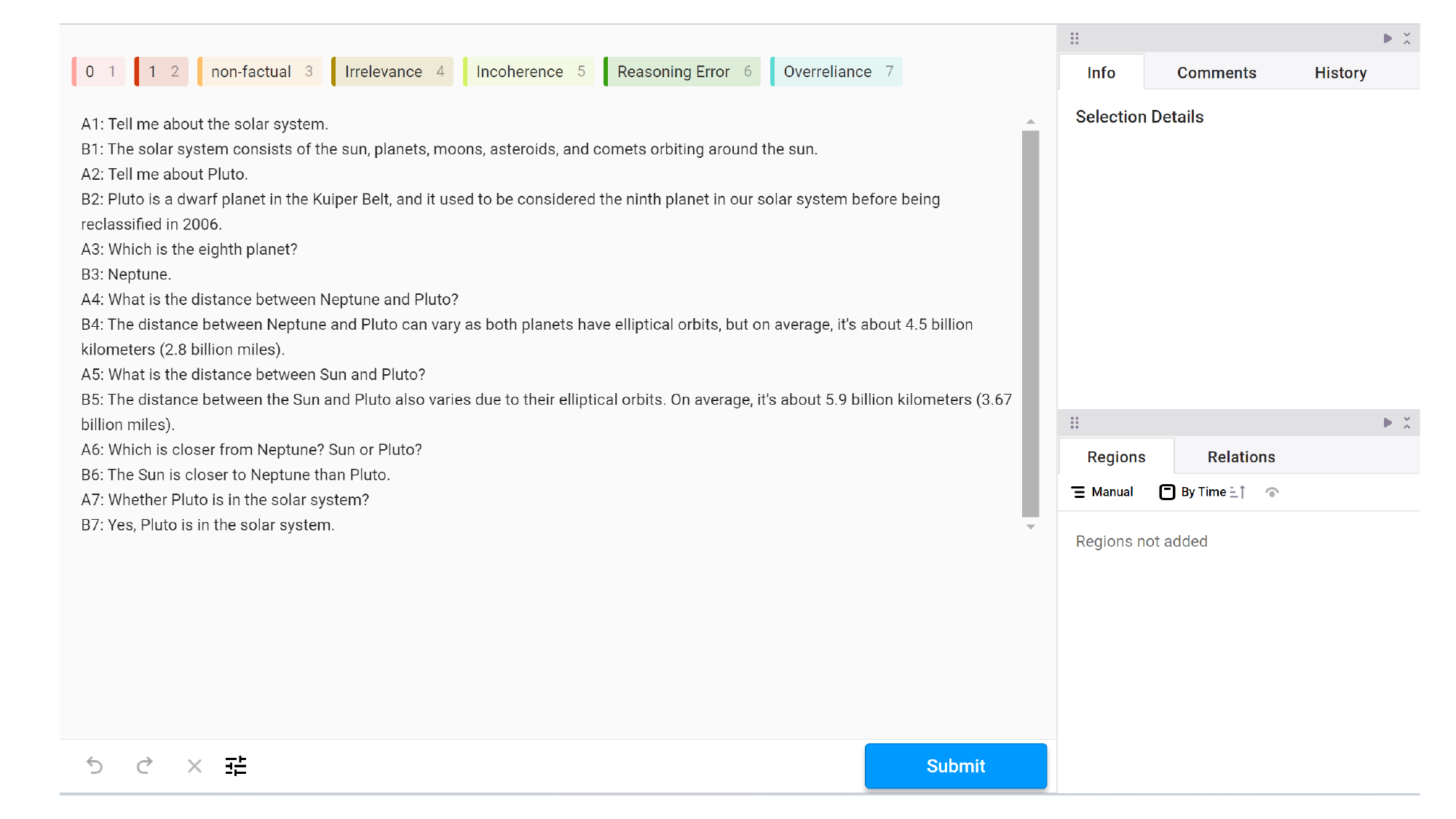}
    \caption{The annotation interface on Label Studio.}
    \label{fig: annotation interface}
\end{figure*}

\paragraph{The Label Platform} We use Label Studio\footnote{https://labelstud.io/} for labeling, which is an online open-source
data labeling platform in the field of artificial intelligence. The annotation interface is depicted in Figure~\ref{fig: annotation interface}.

\subsection{The Application Scope of Hallucination Labels}
\label{app: application scope of hallucination labels}
Annotating the hallucination and its subtypes in this benchmark is a very challenging task. One of the reasons is that some hallucination subtypes in edge cases are defiant to differentiate. Therefore, in the first stage of annotation, we provide detailed definitions for each hallucination subtype. Below are the results of the discussion between the experts and the annotators.

\paragraph{Non-factual} implies that it does not align with facts or introduce elements that do not exist in real life.

\paragraph{Incoherence} means there is a contradiction between one of the speakers and himself in the dialogue context, a contradiction between the two speakers (when both are declarative sentences) in the dialogue context. It also refers to factual and relevant nonsense, contextually inappropriate responses or other inconsistent errors. 

\paragraph{Irrelevance} involves responses that are irrelevant to the dialogue topic or an irrelevance due to misunderstanding the grammar of a question. (Please note that we emphasize the use cases for both interrogative and declarative sentences.)

\paragraph{Overreliance} is that the LLM excessively trust in the correctness of the context, generating serious responses to statements that were inherently wrong or unanswerable (in a declarative sentence).

\paragraph{Reasoning Error} covers all errors within the reasoning dialogue.

\subsection{Baselines and Metrics}
\label{app: baseline}

\textbf{I. Baselines Selected} 

Below is a detailed description of all the baselines we selected.

\paragraph{Random} A straightforward approach that randomly generates a label for each sample.

\paragraph{FaithCritic} \citep{DBLP:journals/tacl/DziriKMZYPR22} is one of the most effective dialog text hallucination classifiers before the era of large language models. Trained on a large-scale dialog corpus, it can output the confidence level for each classification label. Since the model's input includes dialog-related knowledge, we use the retrieved contents as the knowledge during the experiment. 

\paragraph{SelfCheckGPT} \citep{DBLP:conf/emnlp/ManakulLG23} It is a widely used black-box hallucination detection framework. It rephrases the contents to be detected while ensuring the consistency of semantics by LLMs with different temperatures. Furthermore, it calculates the consistency between the original and the rephrased contents using five methods, thereby determining whether there exits hallucination. 
The indices $B$, $N$ and $P$ respectively denote scoring with BERTScore \citep{DBLP:conf/iclr/ZhangKWWA20}, scoring with Natural Language Inference methods \citep{DBLP:conf/iclr/HeGC23} and the direct judgment using prompts.

\paragraph{FOCUS} \citep{DBLP:conf/emnlp/ZhangQGDZZZWF23} is an improved version of SelfCheckGPT. It takes into account the attention scores between entity tokens, enabling more accurate classification of hallucination at both the sentence and paragraph levels.

\paragraph{LLaMa-30B \&\& Vicuna-33B} They are two well-pretrained and widely deployed open-source LLM backbones\footnote{https://huggingface.co/huggyllama/llama-30b, https://huggingface.co/lmsys/vicuna-33b-v1.3} \citep{DBLP:journals/corr/abs-2302-13971,23vicuna}. We provide a specially designed prompt to assist with detection. More details about this prompt are shown in Appendix~\ref{app: the prompt}.

\paragraph{Gemini1.5 PRO}
Gemini1.5 PRO\footnote{https://gemini.google.com/} \citep{DBLP:journals/corr/abs-2312-11805} is the latest version of the language model launched by Google. It inherits the powerful natural language processing capabilities of its predecessor and has made significant improvements in understanding and generating text. We employ the same prompt for binary detection as LLaMa-30B and Vicuna-33B do. We also create a manually prompt to assist with fine-grained recognition in Appendix~\ref{app: the fine-prompt}.

\paragraph{ChatGPT3.5 \&\& GPT4} Both of these models are developed by OpenAI\footnote{https://chat.openai.com/}.
ChatGPT3.5 marks the beginning of the era of large language models and GPT4 is currently the most powerful language model \citep{10113601,openai2023gpt4}. We employ the same prompt as LLaMa-30B and Vicuna-33B do for binary detection. And the same prompt as Gemeni1.5 PRO do is used for fine-grained recognition.
The ChatGPT3.5 version is ChatGPT3.5 (gpt-3.5-turbo)\footnote{https://platform.openai.com/docs/models/gpt-3-5-turbo} and the GPT4 version is GPT4 (gpt-4-turbo) \footnote{https://openai.com/gpt-4}.

\textbf{II. Metrics Calculation} 

Despite we annotating the subtypes of hallucination in the dataset, achieving consistent labels even among humans requires further discussion. Therefore, similar to past hallucination detection efforts, we first focus on a binary classification task of determining the existence of hallucination. Consequently, we utilize binary classification evaluation metrics: Precission, Recall and F1. The positive label for this classification task is set as "Halu", for our main focus is testing the model's ability to recognize hallucination.

As for more fine-grained hallucination-type recognition, We define a correct judgment as one where both the presence of hallucination and the specific subtype of hallucination are accurately identified. For all label types, we use micro-F1 score to quantify the performances of the three classification models.

\subsection{The Prompt Designed for Detection}
\label{app: the prompt}
\begin{figure*}[!t]
    \centering
    \includegraphics[scale=0.5]{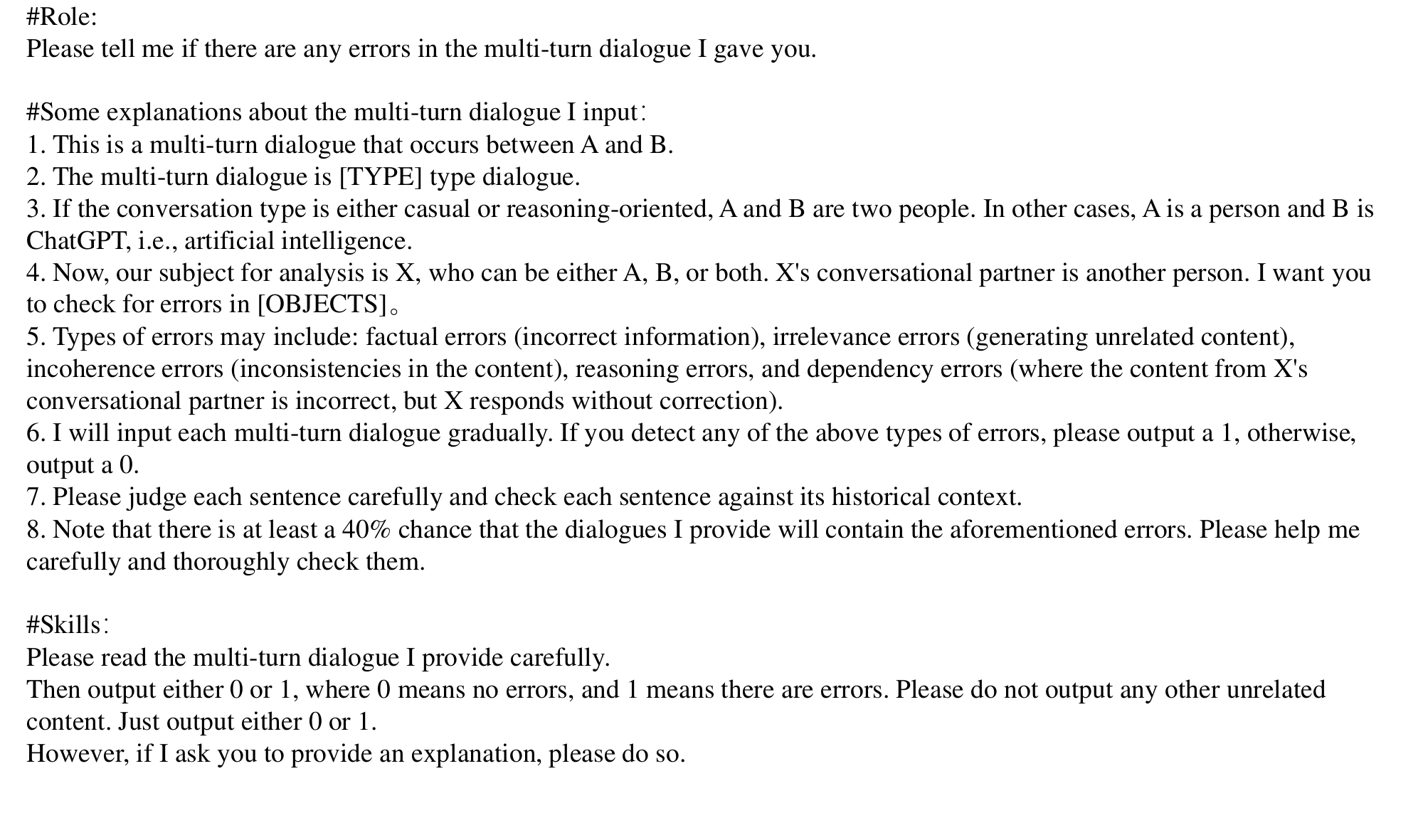}
    \caption{The whole prompt for hallucination detection of the baselines.}
    \label{fig: the prompt}
\end{figure*}

In Figure~\ref{fig: the prompt}, we show the whole prompt specially designed for hallucination detection of the baselines: LLaMa-30B, Vicuna-33B, Gemini1.5 PRO, ChatGPT3.5, and GPT4. 
It is worth noting that due to the poor instruction-following ability and the disorderly output format of the LLaMa and Vicuna models, we conduct experiments in a 1-shot manner. 

Previous works prove the LLMs' ability to follow complex instructions \citep{DBLP:conf/emnlp/ManakulLG23,DBLP:journals/corr/abs-2310-14564,DBLP:journals/corr/abs-2310-00741,mündler2023selfcontradictory,DBLP:conf/emnlp/LiCZNW23,DBLP:journals/corr/abs-2303-04048,DBLP:conf/emnlp/LiuIXWXZ23,DBLP:journals/corr/abs-2401-07103}, including some hallucination tasks. Thus, we reference those prompts that classify hallucination using LLMs, and then we formulate prompts for ours.

\subsection{The Prompt Designed for Fine-grained Recognition}
\label{app: the fine-prompt}
\begin{figure*}[!t]
    \centering
    \includegraphics[scale=0.3]{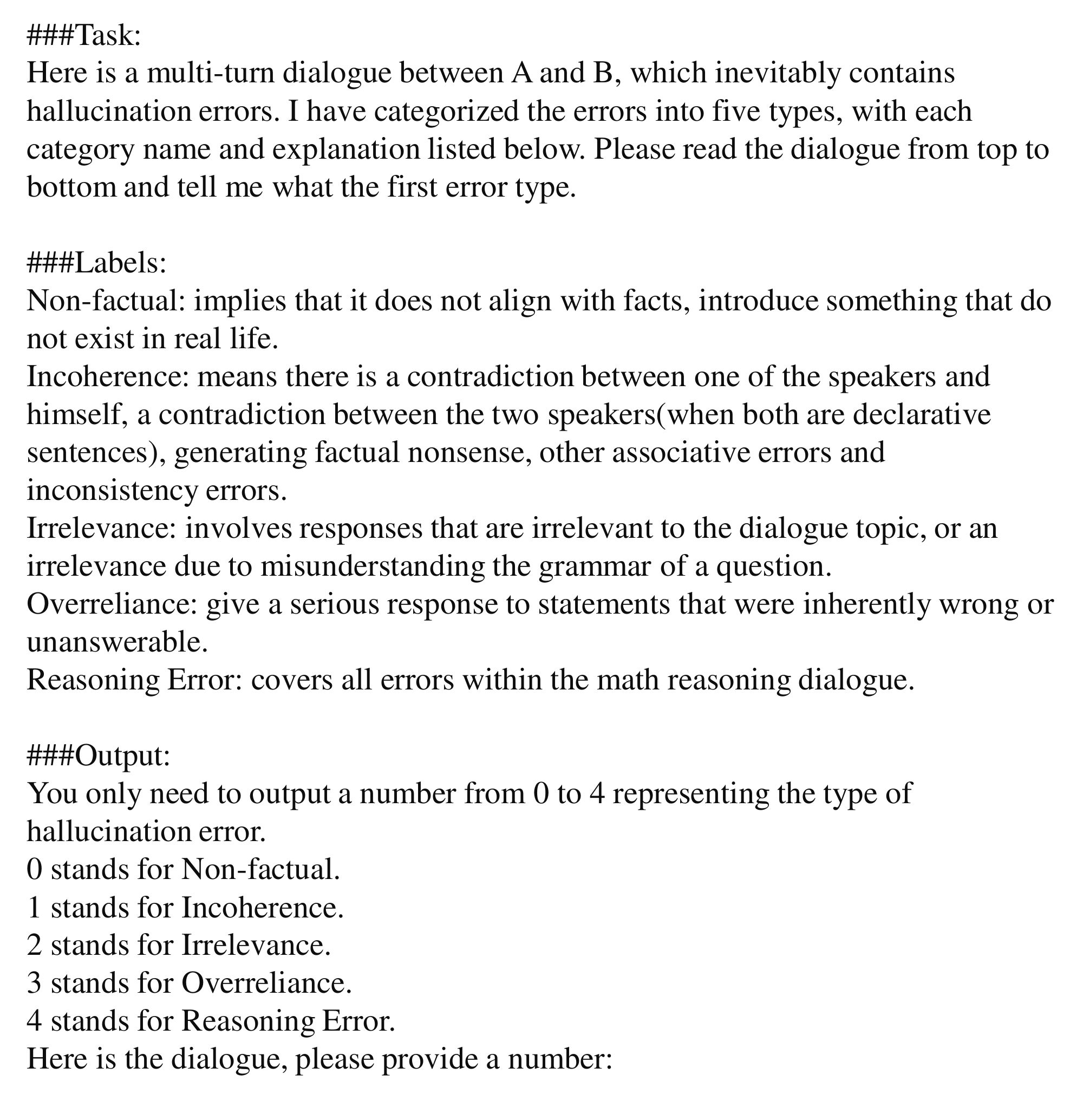}
    \caption{The whole prompt for fine-grained hallucination-type recognition.}
    \label{fig: the fine-prompt}
\end{figure*}

In Figure~\ref{fig: the fine-prompt}, we show the whole prompt manually created for fine-grained hallucination-type recognition of the three closed-source baselines: Gemini1.5 PRO, ChatGPT3.5, and GPT4.

\subsection{The Settings for CoT and Retrieval }
\label{app: CoT and Retrieval}
\begin{figure*}[!t]
    \centering
    \includegraphics[scale=0.45]{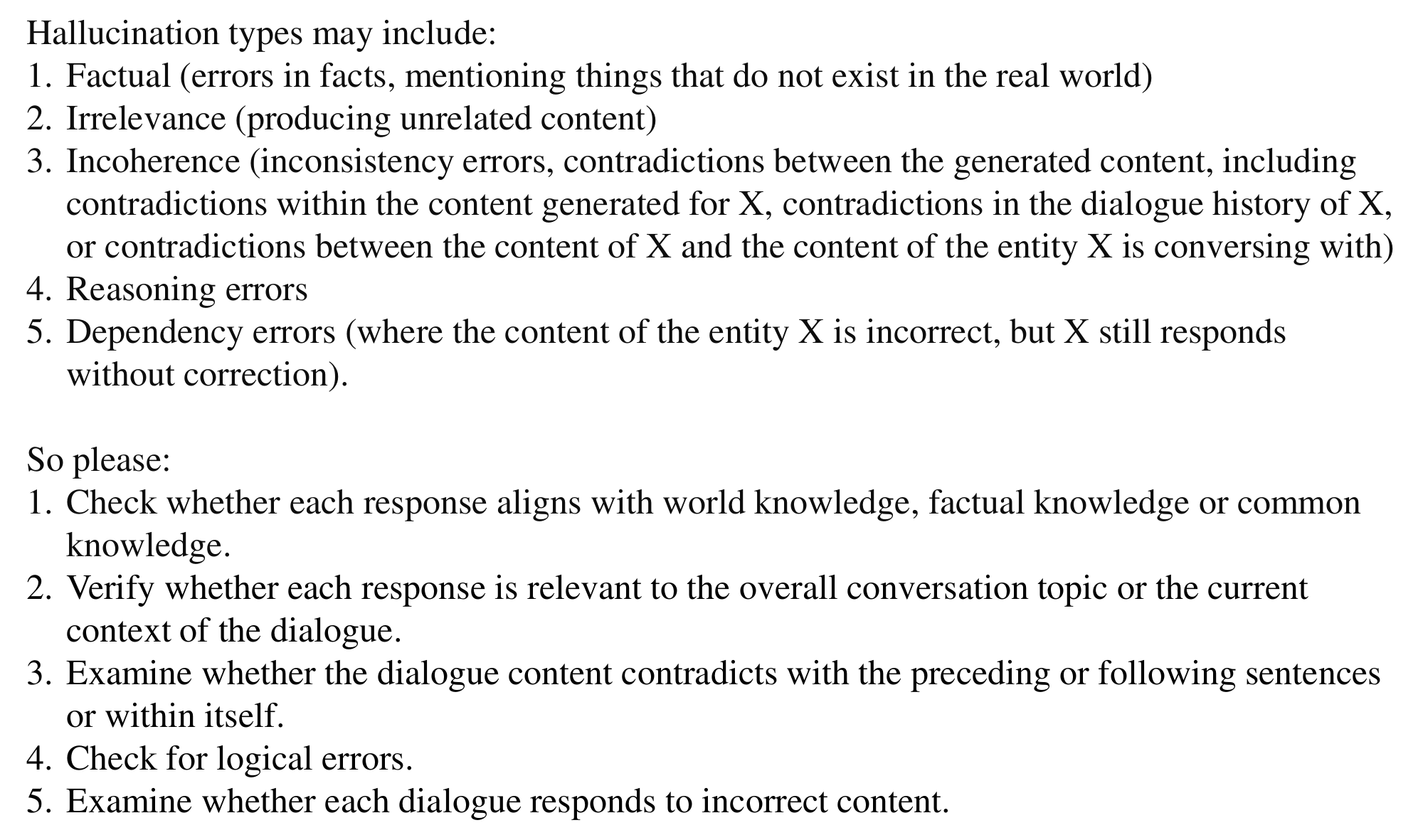}
    \caption{The whole CoT for the four domains of dialogue.}
    \label{fig: the cot}
\end{figure*}

Chain-of-Thought (CoT) \citep{DBLP:journals/corr/abs-2305-15408} describes the organized sequence of logical reasoning that unfolds during thinking. Retrieval \citep{DBLP:journals/corr/abs-2312-10997,DBLP:journals/corr/abs-2403-18243} means retrieving relevant contents from the media to supplement external knowledge for LLMs. 
We employ CoT in all four domains to enhance the performance of ChatGPT3.5 and GPT4. The specific CoT is illustrated in Figure~\ref{fig: the cot}. Even so, only knowledge-grounded and reasoning domains are tested with retrieval via Google\footnote{https://console.cloud.google.com/apis/library}. 
This is because domains of task-oriented and chit-chat mainly involve scenarios related to daily life or virtual worlds, without specific domain knowledge as supplementary information.

\subsection{Future Works}
\label{app: future works}
\paragraph{The necessity of dialogue-level}

The sentence-level, passage-level and dialogue-level hallucination differ in the hallucination types and detection difficulties. We will explain this with the following examples to make it more clear.

Here is a sentence-level hallucination example from the dataset FactCHD \citep{DBLP:journals/corr/abs-2310-12086}. ‘User: Can you tell me which mountain range is longer, the Alps or the Pyrenees? LLMs: The Pyrenees are longer than the Alps.’ In this sentence-level example, the LLM only responds to the user's question with one sentence. We need to assess whether there are hallucination in the single sentence generated by the LLMs.

Here is a passage-level hallucination example from the dataset WikiBio \citep{DBLP:conf/emnlp/ManakulLG23}. "Matthew Aylmer, 1st Baron Aylmer was an Irish soldier and colonial administrator. He was born in Dublin, the son of a barrister, and was educated at Trinity College, Dublin. ... He was buried in Westminster Abbey." This passage-level example is directly generated by the LLM. We need to determine whether the passage with multiple sentences involves hallucination. In this example passage, hallucination occurs in the last sentence which provides unfactual information. Since there are interrelations or dependencies between the sentences in the passage, passage-level hallucination detection is more challenging than the sentence-level one.

The examples of dialogue-level hallucination are shown in Figure 2, which covers four domains and five hallucination types. The differences between it with sentence-level and passage-level hallucination are as follows:
\textbf{First}, more types of hallucination occur frequently in dialogue. One possible reason for most benchmarks merely focusing on detecting factuality hallucination is that they are organized at a sentence or passage level. In this setting, faithfulness hallucinatios (including Incoherence, Irrelevance, and Overreliance) are less likely to occur. In contrast, since dialogue generation requires LLMs to have context coherence \citep{DBLP:conf/aaai/MishraPE23}, track the dialogue state \citep{DBLP:conf/acl/HeckLRVFGLNG23}, possess long-term memory capabilities \citep{DBLP:conf/aaai/ZhongGGYW24}, and have the ability to recognize topic shifts \citep{DBLP:conf/icdar/LinFJCL23}, faithfulness hallucination (including Incoherence, Irrelevance, and Overreliance as described in lines 261-263) occur more frequently in dialogue. In our benchmark, the faithfulness hallucination mainly accounts for Task-oriented dialogue and Chit-Chat dialogue as shown in Figure 4.
\textbf{Second}, it is more challenging to detect hallucination from dialogue than a single sentence or a passage. Since a dialogue contains multiple rounds of interactions that are interdependent, it can not determine whether the current round has hallucination merely based on the current content. For example, the Incoherence hallucination type occurs as the answer is not consistent with the previous context in the task-oriented dialogue in Figure 2. Thus, it needs to analyze the context dependency in the whole dialogue and judge coherence, relevance and reasoning correctness, spanning multiple rounds of interactions for dialogue-level hallucination detection, which is more challenging than sentence-level and passage-level detection.

Therefore, it is necessary to construct a dialogue-level hallucination evaluation benchmark to promote research in LLM studies.

\paragraph{Dialogue-level hallucination detection} is an important work in the future. We propose the first dedicated dialogue-level hallucination detection evaluation benchmark for LLMs and experimental results show that it is a very challenging task. Therefore, combining previous works \citep{DBLP:journals/sigkdd/ChenLYT17,DBLP:journals/air/DeriuROERAC21}, developing methods based on this dataset to achieve a relatively high recognition accuracy is highly valuable.

\paragraph{Dialogue-level hallucination elimination} is an extension task of this work. Most existing hallucination elimination methods primarily focus on sentence-level or passage-level factuality hallucination \citep{DBLP:journals/corr/abs-2401-08358}. 
Hallucination elimination at the dialogue level not only requires models to have much parameter knowledge, but also a long-context memory capabilities, the abilities to recognize changes in topics/roles and logical transitions in the dialogue. These are helpful in addressing faithfulness hallucination. At the same time, improving the accuracy of knowledge in the LLMs' parameters and the reasoning abilities are equally important.

\paragraph{Hallucination snowballing} is the phenomenon that LLMs tend to accumulate hallucination rather than self-correcting during the generation process \citep{DBLP:journals/corr/abs-2309-01219}. 
Some previous works validate this phenomenon \citep{DBLP:journals/corr/abs-2304-13734,DBLP:conf/icde/AngHTH23}. 
In our benchamark, there is a noticeable issue of hallucination snowballing. Through the experimental results, we also display such phenomenon. This is because that LLMs are unable to perform timely self-check during generation, leading to the accumulation of hallucinations in multi-turn dialogues.
Eliminating hallucination snowballing in LLMs is extremely urgent in the future.

\paragraph{The Unanswerability of LLMs} During the annotation process of this dataset, we introduce a hallucination category termed "overreliance", which represents answering unanswerable content \citep{DBLP:journals/corr/abs-2310-11877, DBLP:conf/emnlp/SulemHR21}. 
This phenomenon signifies that LLMs tend to trust the input provided by users. Sometimes, even when there are errors in user input, the LLMs still fail to recognize them. A few past researches explore the related areas and try to find a solution. However, this issue in the application of human-machine interaction and multi-agent scenarios still remains crucial.
\end{document}